\documentclass[letterpaper, 10 pt, conference]{ieeeconf}  % Comment this line out if you need a4paper

\IEEEoverridecommandlockouts                              % This command is only needed if 
\usepackage{graphics} % for pdf, bitmapped graphics files
\usepackage{epsfig} % for postscript graphics files
\usepackage{mathptmx} % assumes new font selection scheme installed
\usepackage{times} % assumes new font selection scheme installed
\usepackage{amsmath} % assumes amsmath package installed
\usepackage{amssymb}  % assumes amsmath package installed
\usepackage{romannum}
\usepackage{multirow}
\usepackage{subfig}
\usepackage{acronym}
\usepackage{xcolor}
\title{\LARGE \bf Hybrid Perception and Equivariant Diffusion for Robust Multi-Node Rebar Tying}

\author{Zhitao Wang$^{1,\dag}$ , Yirong Xiong$^{1,\dag}$ , Roberto Horowitz$^{2}$, Yanke Wang$^{3, *}$, Yuxing Han$^{1, *}$% <-this % stops a space
\thanks{$\dag$ Equal contribution.}% <-this % stops a space
\thanks{* Corresponding authors: Yanke Wang ({\tt\small yankewang@ust.hk}), Yuxing Han ({\tt\small yuxinghan@sz.tsinghua.edu.cn}).}
\thanks{$^{1}$Zhitao Wang, Yirong Xiong, and Yuxing Han are with Tsinghua University, Shenzhen International Graduate School, Shenzhen, China.
        {\tt\small wzt22@mails.tsinghua.edu.cn, xyrout@outlook.com}}%
\thanks{$^{2}$Roberto Horowitz is with Department of Mechanical Engineering, University of California, Berkeley. {\tt\small horowitz@berkeley.edu}}
\thanks{$^{3}$Yanke Wang is with Hong Kong Center for Construction Robotics, The Hong Kong University of Science and Technology, Units 808 to 813 and 815, 8/F, Building 17W, Hong Kong Science Park, Pak Shek Kok, New Territories, Hong Kong, China.}%
\thanks{This article is accepted by The IEEE International Conference on Automation Science and Engineering (CASE) 2025.}
}

\begin{document}

\maketitle
\thispagestyle{empty}
\pagestyle{empty}

%%%%%%%%%%%%%%%%%%%%%%%%%%%%%%%%%%%%%%%%%%%%%%%%%%%%%%%%%%%%%%%%%%%%%%%%%%%%%%%%
\begin{abstract}
Rebar tying is a repetitive but critical task in reinforced concrete construction, typically performed manually at considerable ergonomic risk. Recent advances in robotic manipulation hold the potential to automate the tying process, yet face challenges in accurately estimating tying poses in congested rebar nodes. In this paper, we introduce a hybrid perception and motion planning approach that integrates geometry-based perception with Equivariant Denoising Diffusion on SE(3) (Diffusion-EDFs) to enable robust multi-node rebar tying with minimal training data. Our perception module utilizes density-based clustering (DBSCAN), geometry-based node feature extraction, and principal component analysis (PCA) to segment rebar bars, identify rebar nodes, and estimate orientation vectors for sequential ranking, even in complex, unstructured environments. The motion planner, based on Diffusion-EDFs, is trained on as few as 5–10 demonstrations to generate sequential end-effector poses that optimize collision avoidance and tying efficiency. The proposed system is validated on various rebar meshes, including single-layer, multi-layer, and cluttered configurations, demonstrating high success rates in node detection and accurate sequential tying. Compared with conventional approaches that rely on large datasets or extensive manual parameter tuning, our method achieves robust, efficient, and adaptable multi-node tying while significantly reducing data requirements. This result underscores the potential of hybrid perception and diffusion-driven planning to enhance automation in on-site construction tasks, improving both safety and labor efficiency.

\end{abstract}

%%%%%%%%%%%%%%%%%%%%%%%%%%%%%%%%%%%%%%%%%%%%%%%%%%%%%%%%%%%%%%%%%%%%%%%%%%%%%%%%
\section{INTRODUCTION}

Rebar tying entails fastening intersecting steel bars with wire in reinforced concrete construction. This task is both labor intensive and ergonomically challenging, raising concerns about worker injuries and reduced productivity~\cite{c1}. Various robotic solutions have been proposed to automate rebar tying; however, these systems often require extensive on-site infrastructure (gantry mounts) or lack adaptive perception for cluttered environments~\cite{c2}. Although commercial tying robots, such as TyBOT, can achieve over 1,000 ties per hour~\cite{c3}, they generally operate only on flat decks with minimal obstructions, limiting their adoption.

A key hurdle in automating rebar tying lies in reliable detection of rebar nodes (the intersection points) and the robot’s ability to perform multi-node tying in sequence without perturbing the loosely laid rebars~\cite{c16}. Many solutions rely on large-scale learning approaches, demanding comprehensive datasets to handle the variability in rebar arrangements~\cite{c4,c5}. Additionally, rebar misalignment or multi-layer configurations introduce further complexity, often leading to incorrect tying or collisions~\cite{c6}. To this end, this requires both robust perception and adaptive motion planning that can generalize across different scenes.

\begin{figure}[!t]
    \centering
    %------------------ 第一行 ------------------%
    \subfloat[Step \Romannum{1}]{
        \includegraphics[width=0.33\columnwidth]{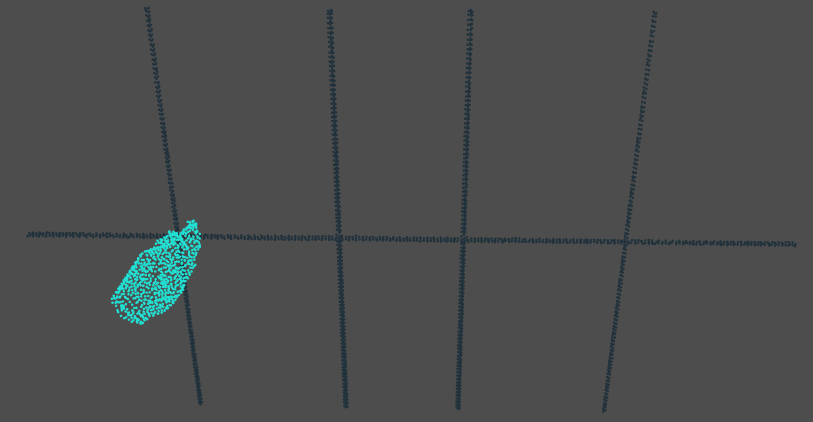}%
        \label{fig:demo1_step1}
    }
    \hspace{0.01\columnwidth}
    \subfloat[Step \Romannum{2}]{
        \includegraphics[width=0.33\columnwidth]{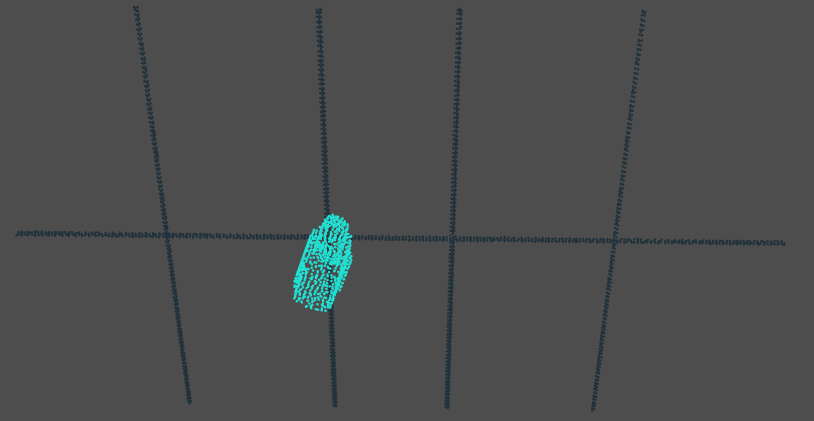}%
        \label{fig:demo1_step2}
    }
    \hspace{0.01\columnwidth}
    \subfloat[Step \Romannum{3}]{
        \includegraphics[width=0.33\columnwidth]{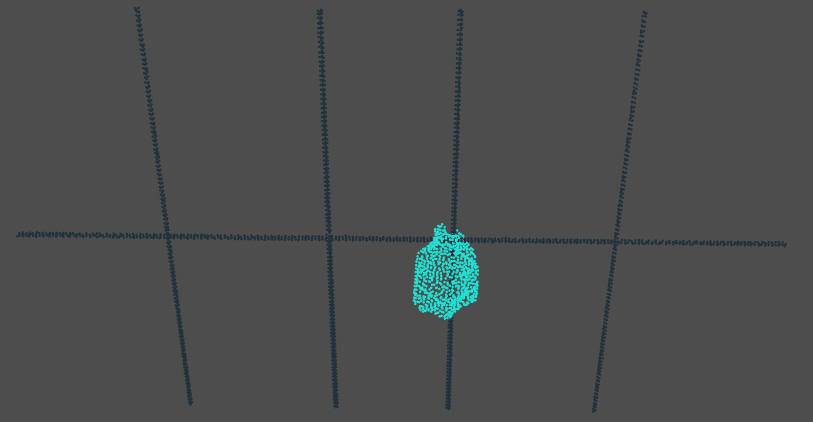}%
        \label{fig:demo1_step3}
    }
    \hspace{0.01\columnwidth}
    \subfloat[Step \Romannum{4}]{
        \includegraphics[width=0.33\columnwidth]{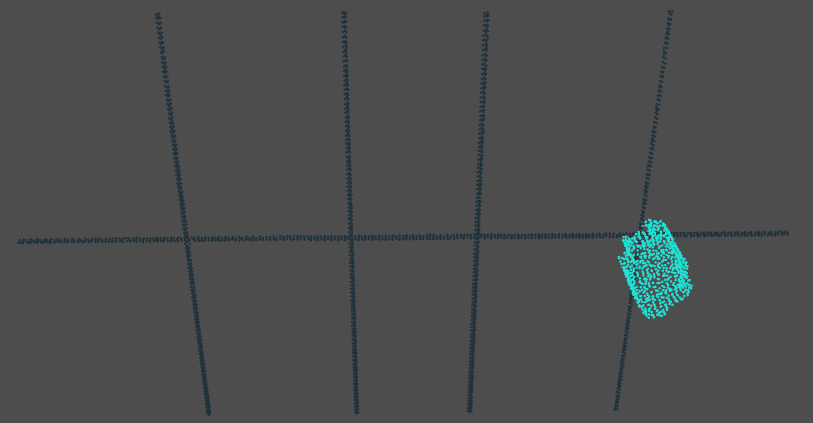}%
        \label{fig:demo1_step4}
    }
    \caption{%
     An example of sequentially tying 4-node rebar set.
    }
    \label{fig:examp_tying}
\end{figure}

We propose an integrated framework that marries a point cloud clustering-based geometry analysis and a Diffusion model with an Equivariant Descriptor Fields (Diffusion-EDFs) for multi-node tying (with an example in Fig. \ref{fig:examp_tying}). On the perception side, density-based clustering (DBSCAN) is exploited to segment rebars from background data, and geometry-based node feature extraction combined with principal component analysis (PCA) is applied to accurately identify intersections for each node and rank them in sequence. These steps effectively localize tie positions even in the presence of overlapping bars or background clutter~\cite{c17}. On the planning side, the Diffusion-EDFs architecture, known to require as few as 5-10 demonstrations for training~\cite{c7}, generates collision-free sequential poses in SE(3). By leveraging symmetry constraints in orientation and translation, the diffusion model efficiently learns how to place the tying tool around multiple intersections in a single pass.

We summarize our contributions as
\begin{itemize}
    \item the design of a multi-node detection algorithm to isolate, cluster, and order rebar intersections for tying,
    \item the integration of Diffusion-EDFs for pose generation with minimal demonstrations, enabling sampling-efficient sequential planning, and
    \item extensive tests and verifications on planar and multi-layer rebar meshes, demonstrating high tying accuracy under real-world conditions.
\end{itemize}
This hybrid approach, combining domain heuristics with an equivariant generative model, significantly reduces the need for large and curated datasets while maintaining robustness in cluttered or unstructured construction settings. The proposed system points to a future where agile and data-efficient robots handle demanding tasks on site, alleviating labor shortages and worker fatigue in reinforced concrete construction~\cite{c18}.

\section{RELATED WORK}

\subsection{Automated Rebar Tying Systems}

Robotic rebar tying has evolved from stationary lab-based platforms to partially on-site automation. Momeni et al.~\cite{c8} automated the production of steel cages in a controlled facility by using a robotic arm, yielding a high throughput but limited generalization outside the lab. Jin et al.~\cite{c9} investigated a crawler tying robot that traversed the rebar mesh; despite its mobility, the robot was prone to displacing unfastened rebars during locomotion. Commercially, TyBOT~\cite{c3} and other gantry-mounted tying robots excel at tying thousands of intersections on large, open decks but face constraints in irregular or congested environments. Overall, robust solutions for multi-layer or cluttered sites remain an active area of research~\cite{c2,c19}.

\subsection{Rebar Node Detection}

A reliable perception pipeline is essential for identifying rebar ties in various rebar configurations. Classical geometric methods segment rebars as linear structures in point clouds and derive crossing points from estimated principal axes~\cite{c6}. Such methods can be highly interpretive and rely on prior manual information. For instance, rebars are usually straight and orthogonal. Vision-based approaches often employ deep neural networks, trained on synthetic or lab-captured images, to identify intersections~\cite{c4,c5}. Although powerful, these detectors can fail under complex real-world conditions (lighting shifts, heavy occlusions) unless the training data are extensive and well-curated~\cite{c10}. For robust on-site performance, recent work combines depth sensing with traditional geometric filters to ensure reliable intersection detection, even with limited training data~\cite{c11,c20}. This aligns with our approach, where a lightweight and geometry-based pipeline is used to isolate rebar nodes across multiple layers or non-uniform meshes.

However, the pipeline combining two-dimensional detection with point cloud post-processing suffers from the difficult alignment and calibration between image and depth information. Additionally, the mentioned methods, especially ~\cite{c11}, only consider the horizontally positioned rebars and the transformation in real world can reduce detection performance. Thus, transformation-invariant methods are in obvious demand.

\subsection{ Diffusion-EDFs for Robotic Planning}
Pose estimation and motion planning in construction robotics often involve enumerating potential collision-free paths, which becomes unwieldy as the scene complexity grows~\cite{c12}. Diffusion models address the challenge by learning a continuous distribution over feasible trajectories. Early work by Ho et al.~\cite{c13} established diffusion models in image generation, inspiring research on applying the same principles to robot control~\cite{c14}. Janner et al.~\cite{c15} demonstrated that a diffusion-based model can outperform conventional trajectory optimization on multi-step tasks. Critically, Ryu et al.~\cite{c7} introduced Diffusion-EDFs, a bi-equivariant model defined on SE(3) that has been successfully applied to tasks such as pick-and-place operations.

Our method integrates geometry-based perception with an equivariant denoising diffusion trajectory planner on SE(3), aiming to tackle the real-world challenges of rebar tying limited data, cluttered meshes, and tight space with minimal data scale. This synergy between geometry-based perception and advanced generative models underscores the potential of data-efficient robotics in construction automation.

As the method on SE(3) ensures the transformation invariance, it naturally fits in the geometry and characters of rebars. Thus, we modify Diffusion-EDF to perform a continuous tying pipeline, which holds its advantages by avoiding complex alignment between image and depth information as well as end-to-end generating the tying pose, with promising application for tying rebars positioned both horizontally and vertically.

\section{Method}
\label{sec:method}

\begin{figure}[htpp]
    \centering
    \includegraphics[width=0.35\textwidth]{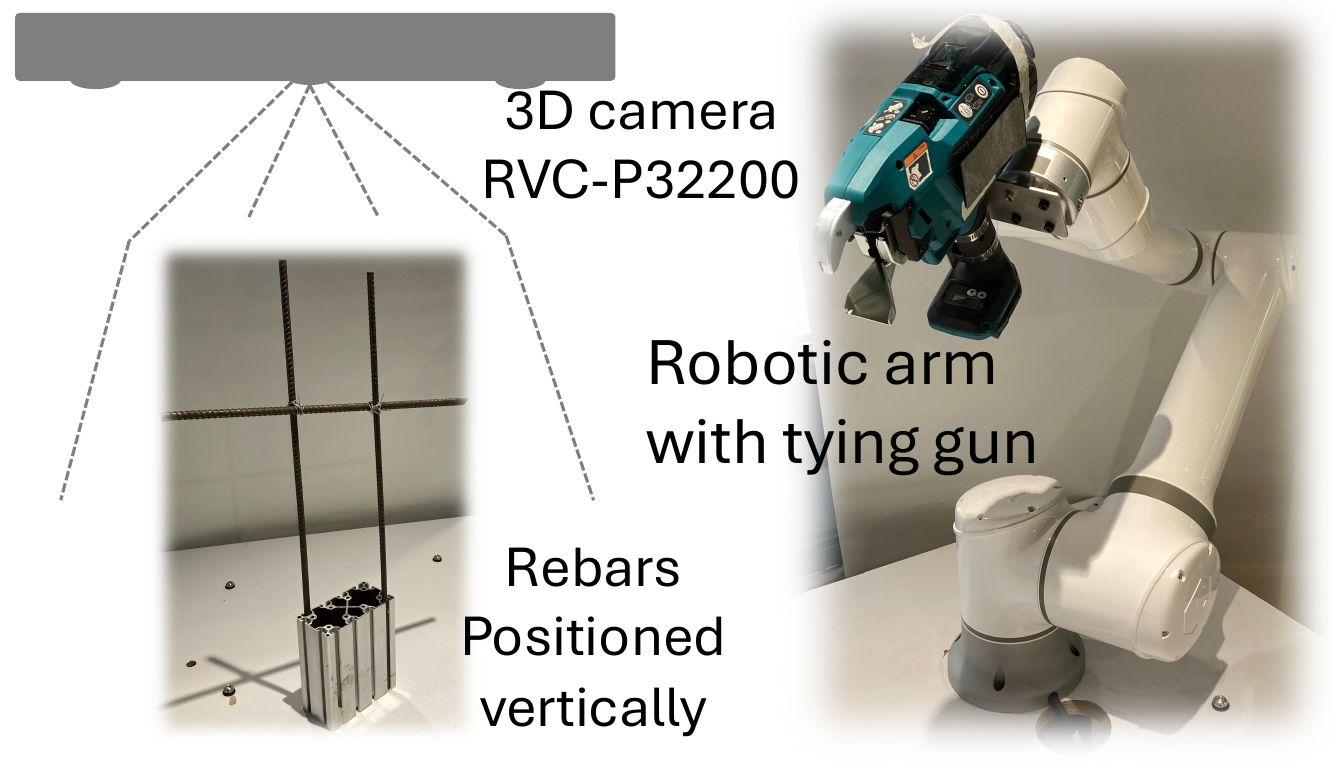} 
    \caption{System design.}
    \label{fig:sys_design}
\end{figure}

\begin{figure}[htbp]
    \centering
    \includegraphics[width=0.25\textwidth]{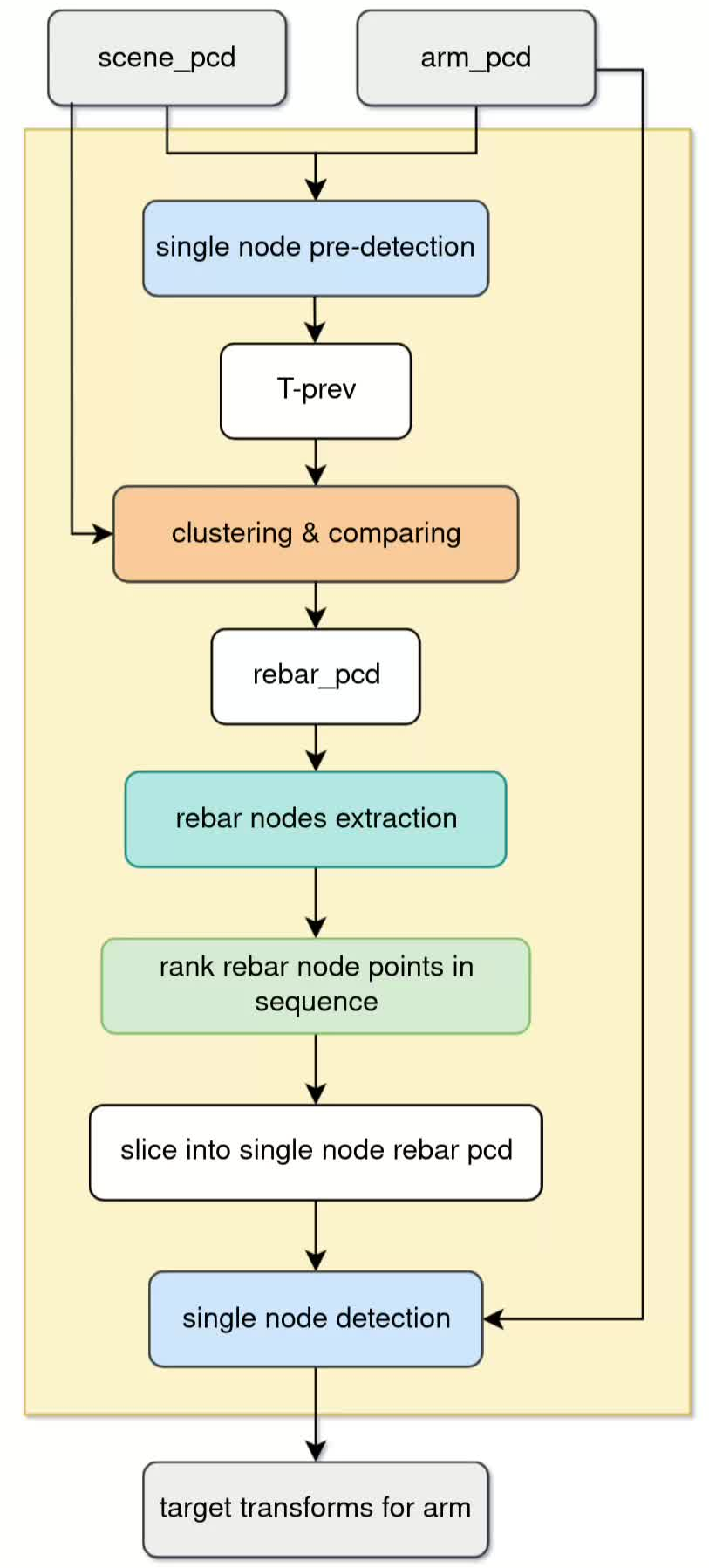} 
    \caption{Overview architecture.}
    \label{fig:architecture_overview}
\end{figure}

In this section, we present the details of our methodology. The system design and overall architecture are illustrated in Fig. \ref{fig:sys_design} and \ref{fig:architecture_overview}. A tying gun is mounted on the end joint of a robotic arm with a 3D camera (RVC-P32200) capturing the colored point cloud of the rebars positioned vertically. Fig. \ref{fig:examp_tying} demonstrates a robotic arm sequentially tying a 4-node rebar structure using our pipeline. Our pipeline operates through the following steps. First, the scene point cloud and the robotic arm point cloud, captured by the 3D camera, are fed into the single-node detector model to compute an initial transformation of the robotic arm, denoted as T-prev. This transformation is then utilized to assist in clustering and isolating the rebar point cloud from the scene point cloud. Next, the extracted rebar point cloud is processed by the rebar node extraction module to identify and extract the nodes, which are subsequently separated and sorted in a specific order. Finally, the single-node rebar point cloud is sliced, and the single-node detection model is applied to compute the transformations for each node individually. Notably, the pipeline is aimed at tying all rebar nodes in the view, and users can customize the tying order. In this work, the nodes are sorted in ascending order along the $x$-, $y$-, and $z$-axes, respectively.
% comprises two principal components. The first component is dedicated to inferring the correct transformations for grasping nodes, utilizing the single node  detection model. The second component handles the sequential grasping of multiple points by segmenting the point cloud of multiple nodes into sorted individual node point clouds.
% \documentclass{article}
% \usepackage{graphicx} % 引入 graphicx 宏包
% \begin{figure}[h] % 图片浮动环境
%     \centering % 图片居中
%     \includegraphics[width=0.5\textwidth]{overview_architecture.png} % 插入图片
%     \caption{overview architecture} % 图片标题
%     \label{fig:overview_architecture} % 图片标签，用于引用
% \end{figure}

\subsection{Single-node Detection Model} 
% Define abbreviations and acronyms the first time they are used in the text, even after they have been defined in the abstract. Abbreviations such as IEEE, SI, MKS, CGS, sc, dc, and rms do not have to be defined. Do not use abbreviations in the title or heads unless they are unavoidable.
Our single-node detection model is designed to identify the pose of individual rebar nodes, built upon the Diffusion-EDFs framework. It is specifically trained to handle tasks involving the tying of a single rebar node. Diffusion-EDFs is a generative model in SE(3) conditioned on observed point clouds, designed for a robotic picking and placing task. The method models the target end-effector pose to normal distribution in SE(3) via a Lie group Stochastic Differential Equation (SDE) as
\begin{equation}
g^{t+dt} = g^t \exp[dW],
\end{equation}
where $dt$ is the diffusion step and $dW$ is the standard Wiener process defined on SE(3) Lie algebra for $g^t$. The model considers scene-pc\textsubscript{d} and gripper point cloud (rebar-tying-pc\textsubscript{d} in our case), so $g^t$ is sampled from $P^t(g_n \mid o_s, o_r)$. The model is trained on a score function $s^t(g^t \mid o_s, o_r) = \nabla \log P^t(g \mid o_s, o_r)$, where $P^t$ is defined via a diffusion process on SE(3) as
\begin{equation}
P^t(g) = \int_{SE(3)} P^{t\mid 0}(g \mid g^0) P^0(g^0) dg^0,
\end{equation}
which diffuses the target pose distribution $P_n^0$ to a standard initial distribution $P_n^t$. The sampling (denoising) process is leveraged via the annealed Langevin MCMC in SE(3) based on the score function, namely
\begin{equation}
g^{t+dt} = g^t \exp \left[ \frac{1}{2} s^t(g^t \mid o_s, o_r) dt + dW \right].
\end{equation}

\subsection{Rebar Point Cloud Detection}
\label{sec:method_clustering}
Although the single-node Diffusion-EDFs model is relatively effective for grasping individual rebar nodes, the similarity among intersections in multi-node point clouds makes the selected node each time essentially random. In practical applications, we need to bind nodes in a specified order; however, Diffusion-EDFs alone cannot guarantee the correct node choice among multiple similar intersections. Therefore, we propose a method to first identify and sort the rebar nodes. Each time the single-node model is applied, only the point cloud segment corresponding to a specific node is fed as an input. This ensures that the robotic arm can sequentially bind the nodes in a predetermined order, as detailed in Section \ref{sec:method_clustering}, Section \ref{sec:method_extract} and Section \ref{sec:method_order}.

The main aim of this part is to separate the rebar point cloud from the background point cloud. To achieve point clustering, we employ the DBSCAN algorithm proposed by Ester et al.~\cite{c21}, a density-based spatial clustering method capable of identifying clusters of any shape based on density conditions. It relies on three key concepts:
\begin{itemize}
\item \textbf{Eps}: The radius defining the neighborhood for identifying core and noise points.
\item \textbf{MinPts}: The minimum number of points required to define a core point.
\item \textbf{Core points}: Points with more than \textbf{MinPts} in their neighborhood.
\end{itemize}

DBSCAN operates by:
\begin{enumerate}
\item identifying core points and forming clusters from their \textbf{Eps} neighborhoods, and
\item expanding clusters by connecting density-reachable points.
\end{enumerate}

The algorithm selects a point $p$ and checks if it satisfies the \textbf{Eps} and \textbf{MinPts} conditions. If $p$ is a core point, its neighborhood forms a cluster; if it is a boundary point, it is marked as noise. By iteratively querying neighborhoods, DBSCAN efficiently mitigates noise and clusters point clouds, effectively distinguishing rebar points from background elements.

\begin{figure}[!t]
    \centering
    %------------------ 第一行 ------------------%
    \subfloat[Input scene point cloud]{
        \includegraphics[width=0.33\columnwidth]{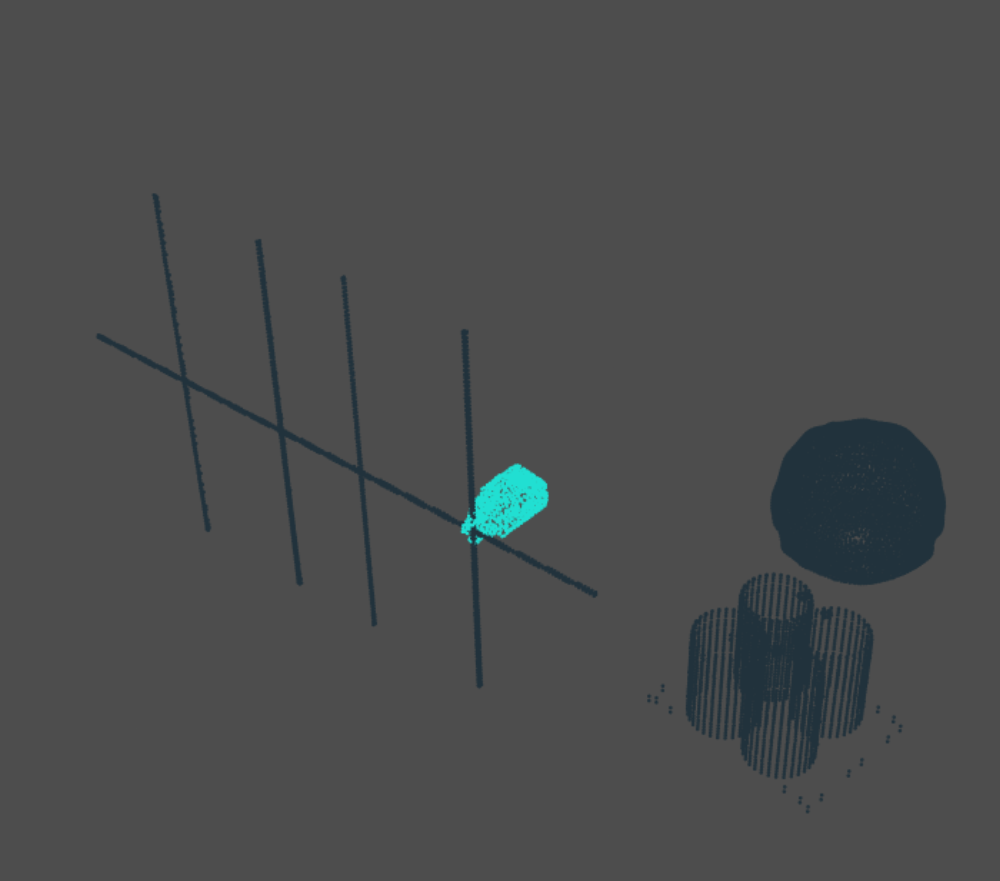}%
        \label{fig:demo1_step1}
    }
    \hspace{0.01\columnwidth}
    \subfloat[Clustering result]{
        \includegraphics[width=0.33\columnwidth]{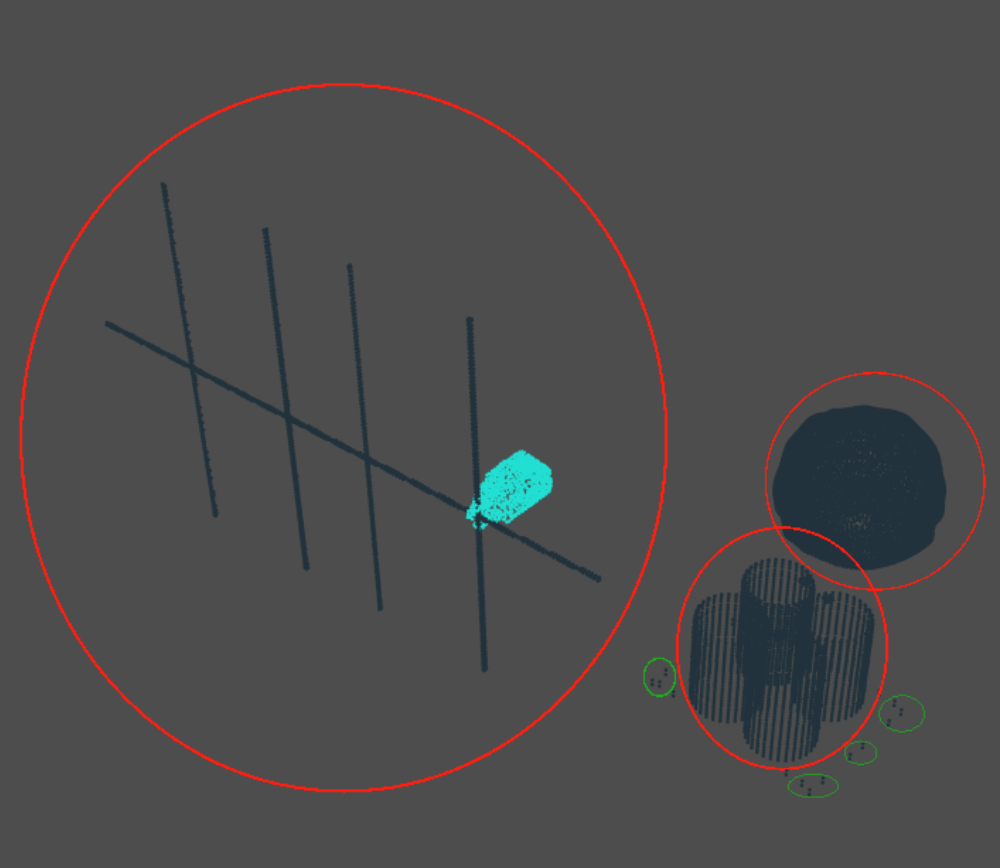}%
        \label{fig:demo1_step2}
    }
    \hspace{0.01\columnwidth}
    \subfloat[Reference point cloud]{
        \includegraphics[width=0.33\columnwidth]{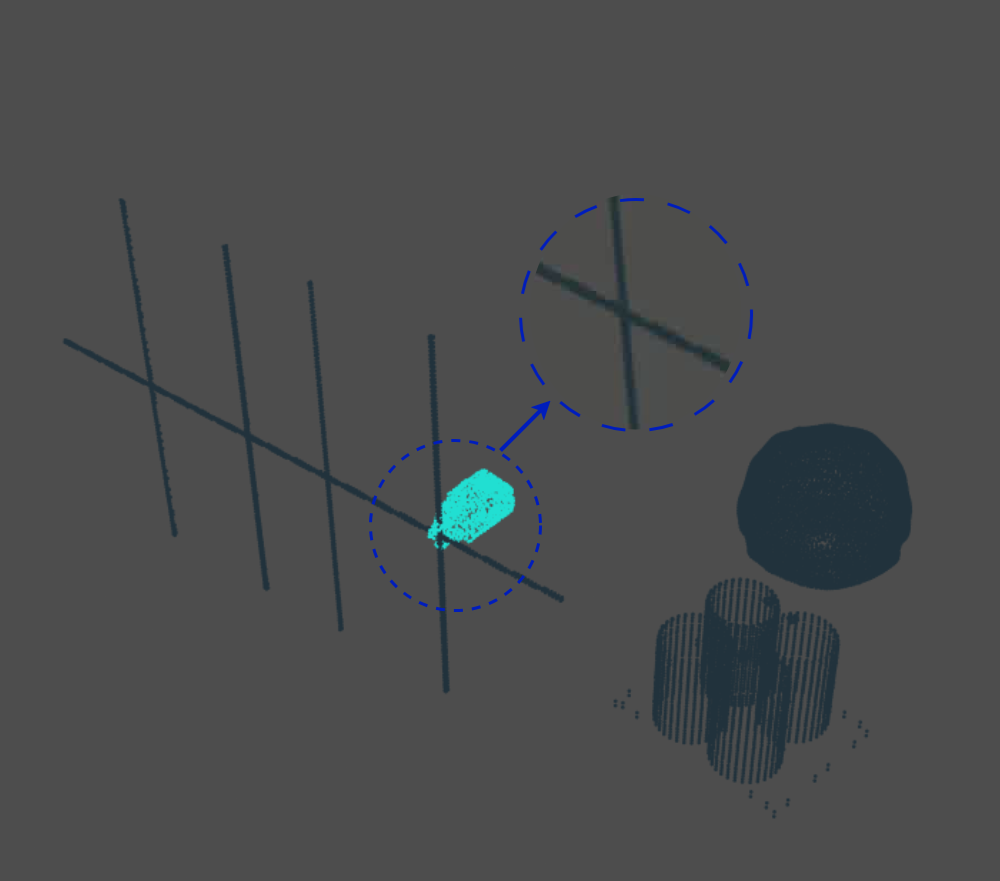}%
        \label{fig:demo1_step3}
    }
    \hspace{0.07\columnwidth}
    \subfloat[Comparing result]{
        \includegraphics[width=0.27\columnwidth]{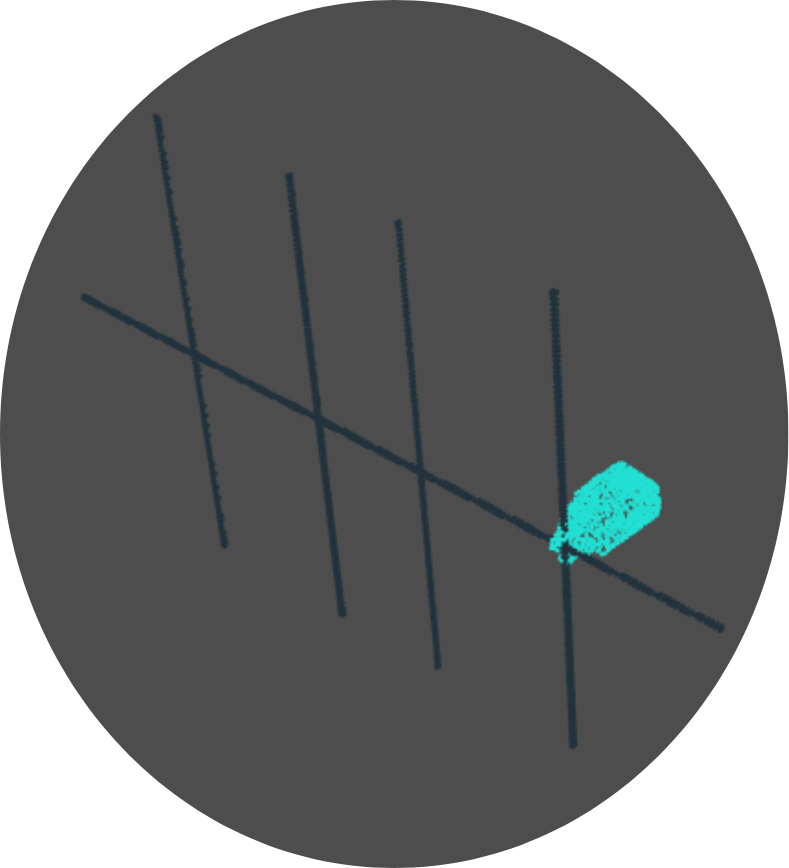}%
        \label{fig:demo1_step4}
    }
    \caption{%
    An example of the clustering process and the selection of the rebar point cloud is illustrated in the figure.
    (a) The original input scene point cloud includes the rebar point cloud and other background obstacles.
    (b) The DBSCAN clustering results show three clusters, highlighted by red circles, along with some noise points marked by green circles.
    (c) A reference point cloud, extracted from the scene near the pre-detected grasp position of a random node, is indicated by a blue circle.
    (d) The final result is obtained by comparing the reference point cloud with the three clusters to identify the target rebar node.
    }
    \label{fig:examp_clustering}
\end{figure}

Subsequently, we employ comparison methods to identify the rebar point cloud among these clusters. Initially, we execute the single-node detection model to obtain T-prev, a transformation matrix designed to guide the robotic arm to grasp a random node within the rebar point cloud. Concurrently, we can calculate the robotic arm's position after applying this transformation, denoted as pose-prev. Next, we extract a small segment of the point cloud near pose-prev to form a reference cluster, which is highly likely to be part of the rebar point cloud rather than other background objects. For each classified target cluster, we calculate the total number of points in the reference cluster that contain points from the target cluster within a search radius. The higher this count, the greater the similarity between the clusters. Ultimately, the target cluster with the highest count is identified as the rebar point cloud. An example of the clustering process and the selection of the rebar point cloud is illustrated in Fig. \ref{fig:examp_clustering}. %\textcolor{red}{I would say, add one image illustraing how all the methods are doung if you have enough time}

% \begin{itemize}

% \item Use either SI (MKS) or CGS as primary units. (SI units are encouraged.) English units may be used as secondary units (in parentheses). An exception would be the use of English units as identifiers in trade, such as Ò3.5-inch disk driveÓ.
% \item Avoid combining SI and CGS units, such as current in amperes and magnetic field in oersteds. This often leads to confusion because equations do not balance dimensionally. If you must use mixed units, clearly state the units for each quantity that you use in an equation.
% \item Do not mix complete spellings and abbreviations of units: ÒWb/m2Ó or Òwebers per square meterÓ, not Òwebers/m2Ó.  Spell out units when they appear in text: Ò. . . a few henriesÓ, not Ò. . . a few HÓ.
% \item Use a zero before decimal points: Ò0.25Ó, not Ò.25Ó. Use Òcm3Ó, not ÒccÓ. (bullet list)

% \end{itemize}

\subsection{Rebar Nodes Extraction}
\label{sec:method_extract}
\begin{figure}[!t]
    \centering
    %------------------ 第一行 ------------------%
    \subfloat[The 16-node rebar point cloud]{
        \includegraphics[width=0.35\columnwidth]{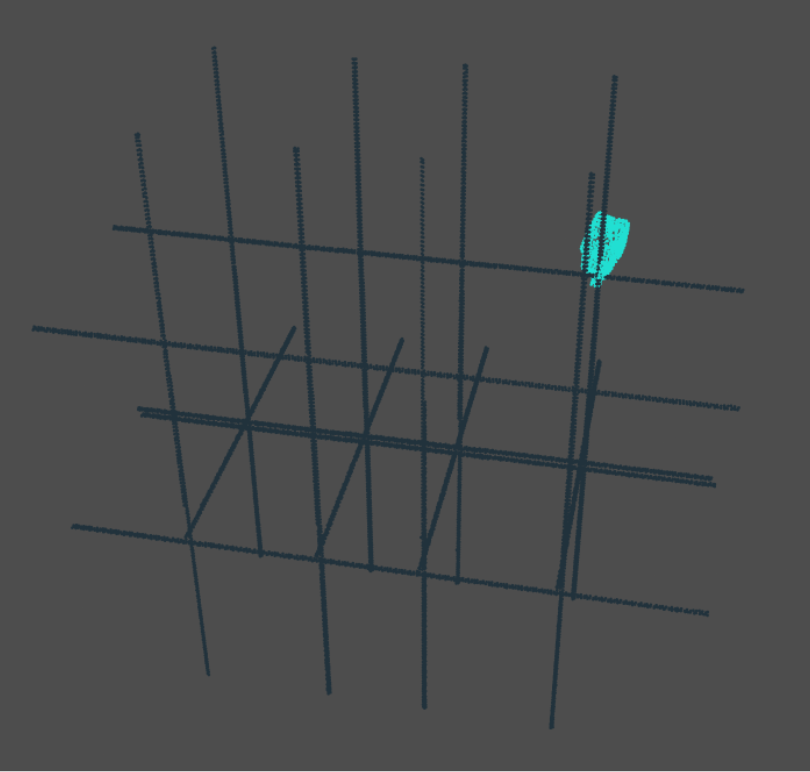}%
        \label{fig:examp_extract_1}
    }
    \hspace{0.01\columnwidth}
    \subfloat[The results visualization of (a)]{
        \includegraphics[width=0.4\columnwidth]{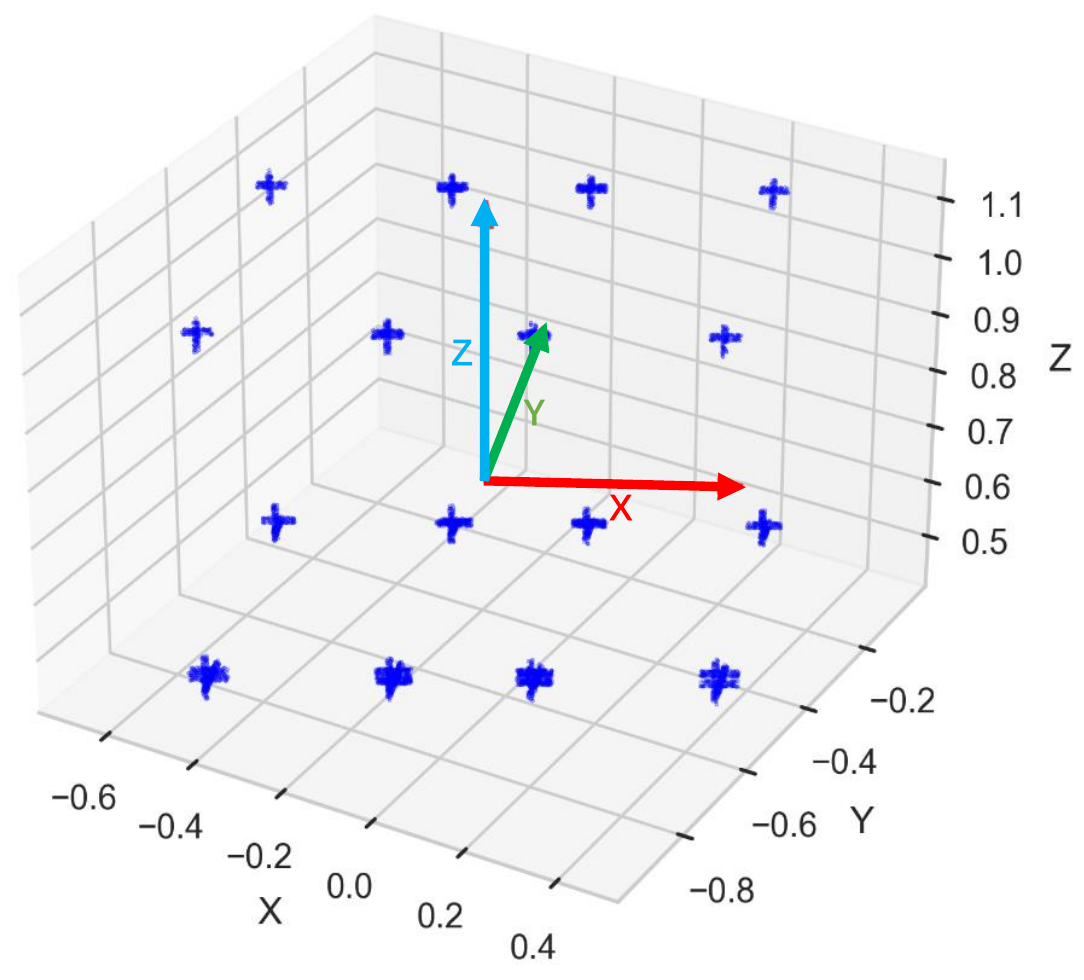}%
        \label{fig:examp_extract_2}
    }

    %------------------ 换行 ------------------%
    \vspace{-1em}

    %------------------ 第二行 ------------------%
    \subfloat[The 16-node rebar point cloud with adjacent obstacles]{
        \includegraphics[width=0.35\columnwidth]{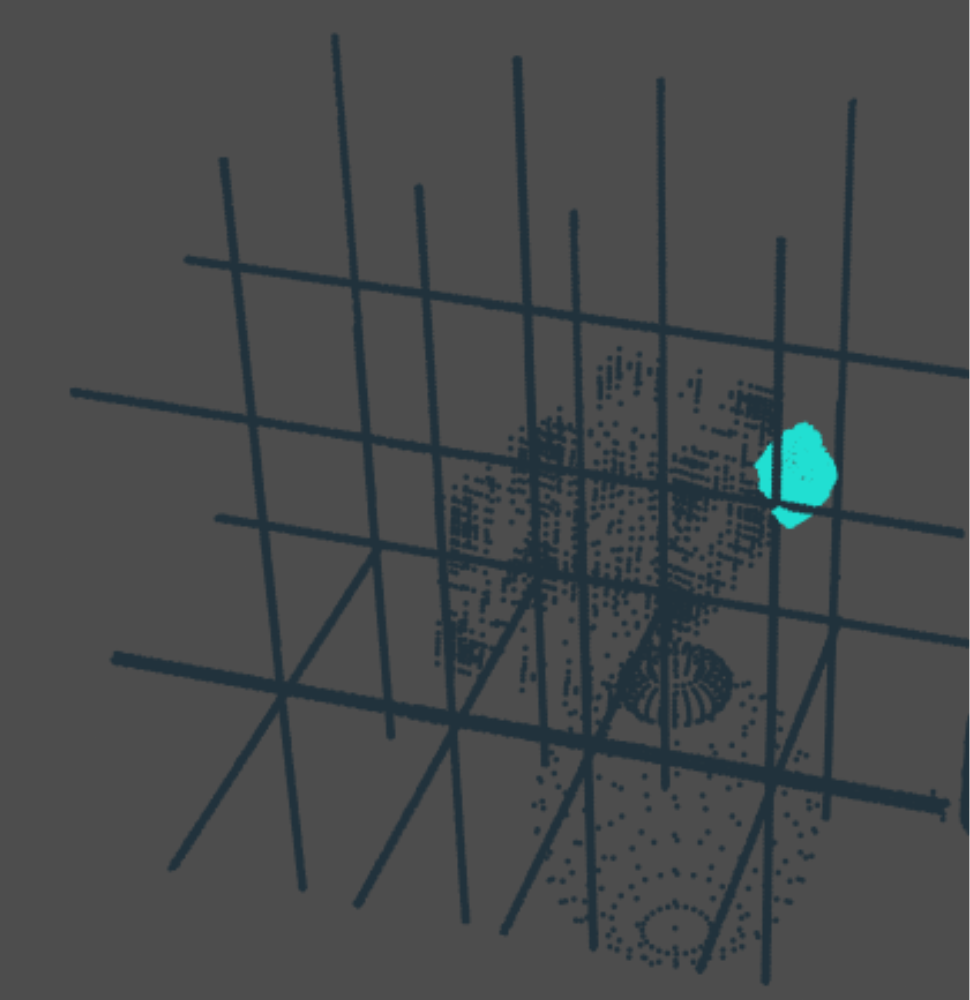}%
        \label{fig:examp_extract_3}
    }
    \hspace{0.01\columnwidth}
    \subfloat[The results visualization of (c)]{
        \includegraphics[width=0.4\columnwidth]{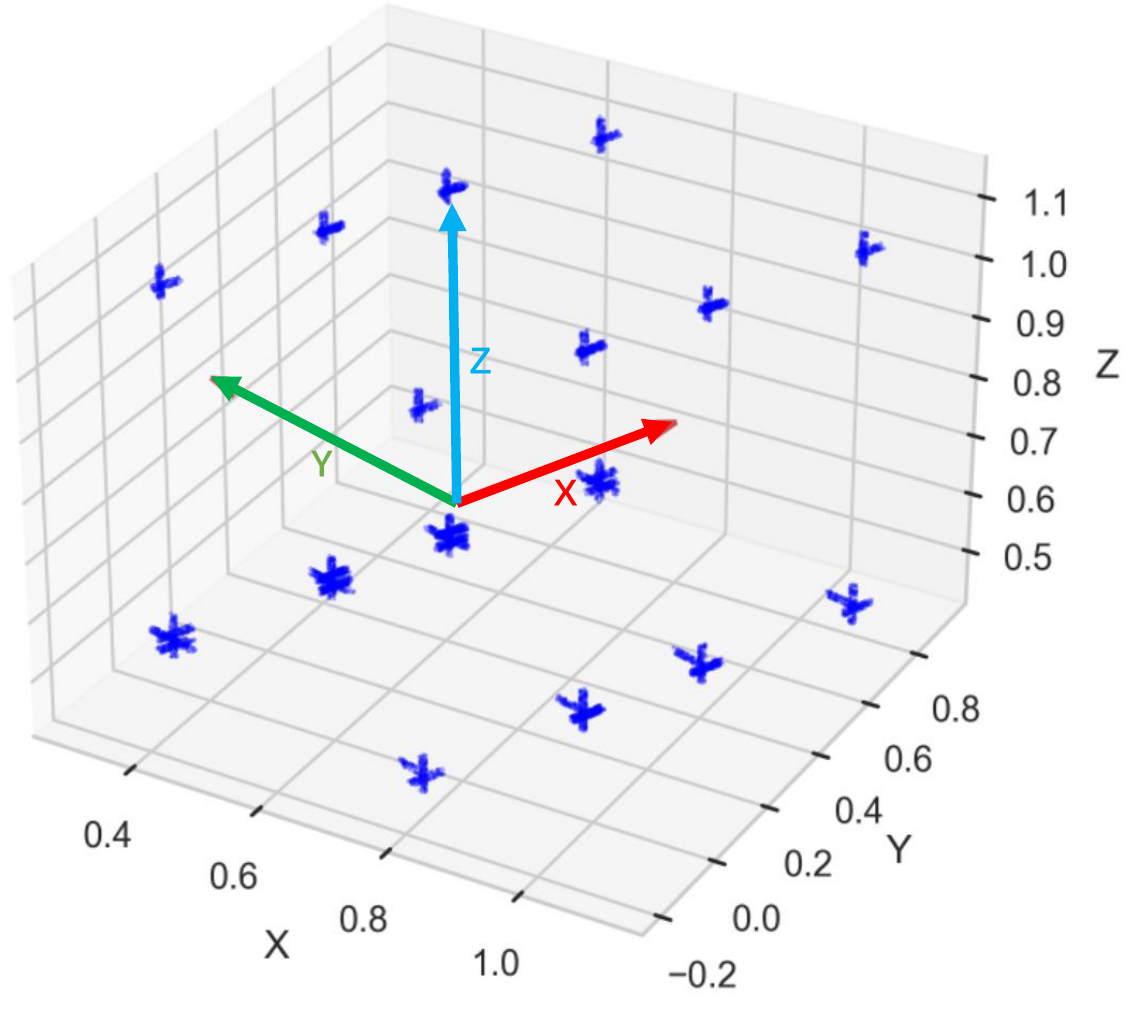}%
        \label{fig:examp_extract_4}
    }
    \caption{%
    An example illustrating the rebar node extraction process and the estimation of direction vectors: (a) and (b) depict the scenario without adjacent obstacles, while (c) and (d) demonstrate the ability to filter out nearby adjacent obstacles that cannot be clustered by DBSCAN.
    }
    \label{fig:examp_extract}
\end{figure}

To precisely locate the nodes within the rebar point clouds, we capitalize on the unique geometric features of the rebar nodes. This approach enables us to efficiently extract points in the vicinity of the nodes while significantly mitigating the interference from background noise. Feature matching is based on orthogonality. Near the locations of the rebar nodes, the point clouds tend to form orthogonal lines. In contrast, near noise-background objects and other similar areas, the point clouds tend to be distributed irregularly. Utilizing this feature, we design an orthogonal feature filter to sort the points near the rebar nodes. The adjustable parameters include:
\begin{itemize}
    \item \textbf{$r_{eps}$}: The radius defining the neighborhood used to determine core and noise points.
    \item $\textbf{$R_{res}$}$: The threshold for determining whether two vectors are approximately perpendicular, which is expected to be set to a value close to 0.
    % The maximum absolute value of the cosine of the angle when judging two vectors to be approximately perpendicular.
    \item $\textbf{$P_{res}$}$: The threshold for judging whether two vectors are approximately parallel, which is expected to be set to a value close to 1.0.
    % The minimum absolute value of the cosine of the angle when judging two vectors to be approximately parallel.
\end{itemize}
The algorithm follows five steps:
\begin{enumerate}
    \item Constructing a k-d tree and searching for neighbors within the radius $r_{eps}$ for each point;
    \item For each point $p$ and its neighbors $N_1$, $N_2$, ..., $N_k$, the set of vectors from $p$ to its neighbors can be represented as $\{v_1, v_2,...v_k\}$, where each vector $v_i$ is defined as $v_i = N_i - p$;
    \item The vectors $\{v_1, v_2,...v_k\}$ are randomly divided into two groups, representing as $\boldsymbol{a} = [a_1, a_2, ..., a_{k/2}]^\top$ and $\boldsymbol{b} = [b_1, b_2, ..., b_{k/2}]^\top$. Then we calculate their dot product of each element as: $\boldsymbol{D} = [a_1\cdot b_1, a_2 \cdot b_2, ..., a_{k/2} \cdot b_{k/2}]^\top$.
    % $\boldsymbol{a} \cdot \boldsymbol{b} = \sum_{i=1}^{k/2} a_i*b_i$. 
    In the two-dimensional case, if the neighboring points are uniformly distributed in an ideal cross shape centered at $p$, due to symmetry, there is a $1/2$ probability that a point lies on the vertical axis and a $1/2$ probability that it lies on the horizontal axis. Consequently, the absolute value of the dot product between any two such vectors can only be 1 or 0, each with a probability of $1/2$. In the three-dimensional case, the probability that a vector is parallel to each of the three mutually orthogonal axes is 1/3. The absolute value of the dot product between two such random vectors has a probability of $P_1=\sum_{i=1}^{3} 1/3 \times 1/3=1/3$ of being 1 and a probability of $P_0=1-P_1=2/3$ of being 0. With consideration of both two-dimensional and three-dimensional cases, the ideal threshold is set to the minimum value.
    \item Given that rebars have a certain width and that actual point cloud contains some noise, the distribution of the actual point cloud cannot lie on an ideal straight line. The absolute dot production probability of being 1 and 0 are both slightly shifted, so we set as adjustable parameters $R_{res}$ and $P_{res}$. The mask is represented as
    \begin{equation}
    M1= len(len(D[ D<R_{res}])\geq len(D) \times 1/2
    \end{equation}
    \begin{equation}
    M2= len(len(D[ D>P_{res}])\geq len(D) \times 1/3
    \end{equation}
    \begin{equation}
   M = M1 \& M2
    \end{equation}
    \item The resulting mask is applied to filter each point, retaining only those that satisfy the condition. For instance, the blue points in Fig. \ref{fig:examp_extract_2} and Fig. \ref{fig:examp_extract_4} respectively illustrate the filtered results of Fig. \ref{fig:examp_extract_1} and Fig. \ref{fig:examp_extract_3}.
\end{enumerate}
For the filtered points, DBSCAN clustering is employed again to separate the point sets at different nodes.
% Then we use another geometric feature to filter again. 
% The equations are an exception to the prescribed specifications of this template. You will need to determine whether or not your equation should be typed using either the Times New Roman or the Symbol font (please no other font). To create multileveled equations, it may be necessary to treat the equation as a graphic and insert it into the text after your paper is styled. Number equations consecutively. Equation numbers, within parentheses, are to position flush right, as in (1), using a right tab stop. To make your equations more compact, you may use the solidus ( / ), the exp function, or appropriate exponents. Italicize Roman symbols for quantities and variables, but not Greek symbols. Use a long dash rather than a hyphen for a minus sign. Punctuate equations with commas or periods when they are part of a sentence, as in

% $$
% \alpha + \beta = \chi \eqno{(1)}
% $$

% Note that the equation is centered using a center tab stop. Be sure that the symbols in your equation have been defined before or immediately following the equation. Use Ò(1)Ó, not ÒEq. (1)Ó or Òequation (1)Ó, except at the beginning of a sentence: ÒEquation (1) is . . .Ó

\subsection{Ordering the Nodes}
\label{sec:method_order}
In Section \ref{sec:method_extract}, we have obtained separate point clusters near different nodes, and we can calculate the mean value of each cluster to determine the approximate position of each node.
To sort them in a certain order, we establish a coordinate system and perform a projection. The direction vectors of the rebar point clouds are estimated by using a Principal Component Analysis (PCA)~\cite{c22}, which is a statistical method to reduce the dimensionality of data while retaining the most significant features. Given that the rebar point cloud data exhibit three primary orthogonal directional distributions, these directions can be effectively estimated through the PCA with the following steps,
\begin{enumerate}
    \item Given a point cloud $P = \{p_1,p_2,...,p_N\}$, where each point $p_i=[x_i, y_i, z_i]^\top$, calculate the mean of the points as $\overline{p} = \frac{1}{N} \sum_{i=1}^{N}p_i$.  % (p_i - \bar{p})(p_i - \bar{p})^T$
    \item Compute the covariance matrix $C$ by
     $C=\frac{1}{N} \sum_{i=1}^{N}(p_i - \bar{p})(p_i - \bar{p})^\top$.
    \item Perform eigenvalue decomposition on the covariance matrix $C$, yielding the eigenvectors and eigenvalues. The eigenvectors correspond to the principal axes of the point cloud, likely aligned with the direction vectors.
    % \item Establish a coordinate system. The average of the rebar points is defined as the origin of the coordinate system, and three coordinate vectors serve as the three axis direction vectors of the coordinate system. The vector closest to the vertical axis of the rebar is selected as the z-axis vector. Utilize the direction of the pose-prev calculated in Section \ref{sec:method_clustering}, relative to the rebar plane to ascertain the y-axis \textcolor{red}{?}. The cross product of these two vectors gives the x-axis.
    \item A coordinate system is established by defining the average of the rebar points as the origin. Three direction vectors are used to determine the axes of the coordinate system. The vector closest to the vertical axis of the rebar is selected as the $z$-axis. The y-axis is determined by identifying the direction vector most aligned with the direction from pose-prev (calculated in Section \ref{sec:method_clustering}) to the mean point of the rebar point cloud. Finally, the $x$-axis is derived by computing the cross product of the $z$-axis and $y$-axis vectors.
\end{enumerate}
First, we construct a coordinate system based on the entire rebar point cloud. To enhance accuracy and reduce the influence of noise or interference from connected background objects, we further refine this process by applying the same steps to the point cloud of each individual rebar node crop. The mean of the coordinate systems derived from these individual crops is then computed to establish a more precise and robust coordinate system. Subsequently, each rebar node is projected along the three coordinate axes within this refined system, and the points are sequentially sorted along the negative $y$-axis, positive $z$-axis, and positive $x$-axis directions. For example, the red arrowed lines in Fig. \ref{fig:examp_extract_2} and Fig. \ref{fig:examp_extract_4} respectively illustrate the axis vector estimation of Fig. \ref{fig:examp_extract_1} and Fig. \ref{fig:examp_extract_3}.

\section{Experiment}

% Use this sample document as your LaTeX source file to create your document. Save this file as {\bf root.tex}. You have to make sure to use the cls file that came with this distribution. If you use a different style file, you cannot expect to get required margins. Note also that when you are creating your out PDF file, the source file is only part of the equation. {\it Your \TeX\ $\rightarrow$ PDF filter determines the output file size. Even if you make all the specifications to output a letter file in the source - if your filter is set to produce A4, you will only get A4 output. }

% It is impossible to account for all possible situation, one would encounter using \TeX. If you are using multiple \TeX\ files you must make sure that the ``MAIN`` source file is called root.tex - this is particularly important if your conference is using PaperPlaza's built in \TeX\ to PDF conversion tool.
% In this section, we designed three groups of experiment to illustrate  
\subsection{Experiment Setup}
In this section, we gather datasets through simulations that encompass a variety of rebar tying scenarios. The experimental process is divided into two main phases. First, we evaluate the robustness of our algorithm. Next, we define two key performance metrics, e.g., success rate and prediction error, and apply them to three representative datasets for assessing the efficacy of the proposed algorithm. All experiments are conducted on a single NVIDIA GeForce RTX 3090 GPU. To conduct a thorough evaluation, we have tested the pipeline on multiple datasets across various scenarios, with specific settings detailed in Table \ref{tab:datasets} and some example visualizations provided in Fig. \ref{fig:demo_visual}.

% \begin{itemize}
%     \item \textbf{Train \Romannum{1}}: 10 demos training set, comprising 1-node rebar tying scenarios, with clear backgrounds
%     \item \textbf{Train \Romannum{2}}: 10 demos training set, comprising 1-node rebar tying scenarios, involving 2 background obstacles
%     \item \textbf{Test \Romannum{1}}: 50 demos testing set, comprising 4-node rebar tying scenarios, involving 4 background obstacles
%     \item \textbf{Test \Romannum{2}}: 50 demos testing set, comprising 8-node rebar tying scenarios, involving 4 background obstacles
%     \item \textbf{Test \Romannum{3}}: 50 demos testing set, comprising 16-node rebar tying scenarios, involving 4 background obstacles
%     \item \textbf{Test \Romannum{4}}: 100 demos testing set, comprising nodes rebar tying scenarios, involving 0 background obstacles
%     \item \textbf{Test \Romannum{5}}: 100 demos testing set, comprising nodes number randomly rebar tying scenarios, involving 4 background obstacles
% \end{itemize}
\begin{table}[h]
\caption{Statistics of the datasets}
\label{tab:datasets}
\begin{center}
\begin{tabular}{|p{40pt}|p{25pt}|p{130pt}|}
\hline
Symbol & Demos & Conditions \\
\hline
\textbf{Train \Romannum{1}} & 10 & 1-node, BG (2 Ob), noise (-)  \\
% \ & & & \\
\textbf{Train \Romannum{2}} & 10 & 1-node, BG (-), noise (-)  \\
\hline
\textbf{Test \Romannum{1}} & 50 & 4-node, BG (-), noise (-)  \\
\textbf{Test \Romannum{2}} & 50 & 4-node, BG (4 Ob), noise (-)  \\
\textbf{Test \Romannum{3}} & 50 & 4-node, BG (-), noise ([0,0.5])  \\
% - & voxel size & 0.01\,m & \multirow{3}{*}{preprocessing} \\
% $(\sigma_c, \rho_c)$ & color jitter & (0.03, 0.35) &  \\
% $(\sigma_p, \rho_p)$ & position jitter & (0.003\,m, 0.35) &  \\
\hline
\textbf{Test \Romannum{4}} & 50 & 8-node, BG (-), noise (-)  \\
\textbf{Test \Romannum{5}} & 50 & 8-node, BG (4 Ob), noise (-)  \\
\textbf{Test \Romannum{6}} & 50 & 8-node, BG (-), noise ([0,0.5])  \\
\hline
\textbf{Test \Romannum{7}} & 50 & 16-node, BG (-), noise (-)  \\
\textbf{Test \Romannum{8}} & 50 & 16-node, BG (4 Ob), noise (-)  \\
\textbf{Test \Romannum{9}} & 50 & 16-node, BG (-), noise ([0,0.5])  \\
% $\gamma$ & regularization term in \eqref{equ:D_g} & 0.1 & evaluation \\
\hline
\textbf{Test \Romannum{10}} & 50 & 32-node, BG (-), noise (-)  \\
\textbf{Test \Romannum{11}} & 50 & 36-node, BG (-), noise (-)  \\
% \textbf{Test \Romannum{11}} & 50 & 16-node,BG(4 Ob),noise(-)  \\
% \textbf{Test \Romannum{12}} & 50 & 16-node,BG(-),noise([0,0.5])  \\
\hline
\multicolumn{3}{p{225pt}}{Demos: Demonstrations} \\
\multicolumn{3}{p{225pt}}{BG: background, noise: Gaussian noise, obstacles: Ob} \\
\multicolumn{3}{p{225pt}}{-: not considered (clear)}
\end{tabular}%
\end{center}
\end{table}

\begin{figure}[!t]
    \centering
    %------------------ 第一行 ------------------%
    \subfloat[1-node demo, on Train \Romannum{2}]{
        \includegraphics[width=0.26\columnwidth]{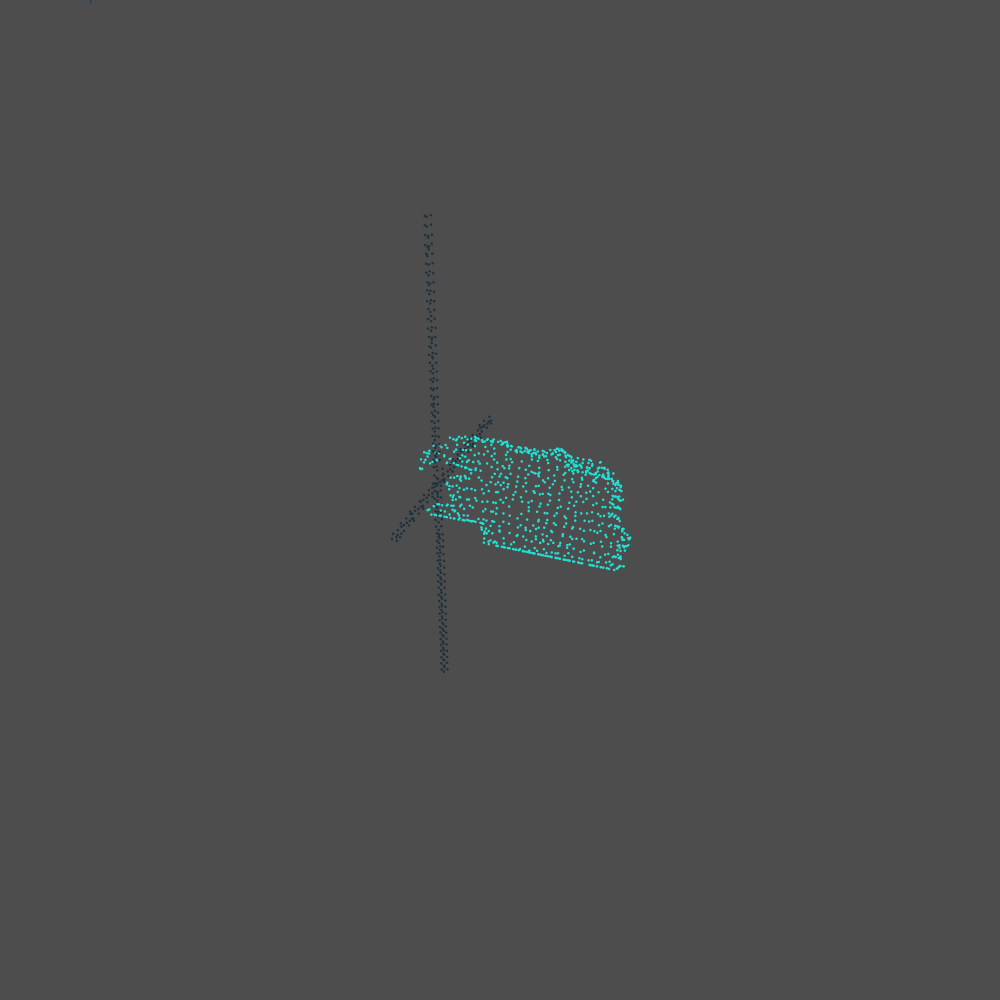}%
        \label{fig:demo1_step1}
    }
    \hspace{0.01\columnwidth}
    \subfloat[4-node demo, on Test \Romannum{1}]{
        \includegraphics[width=0.26\columnwidth]{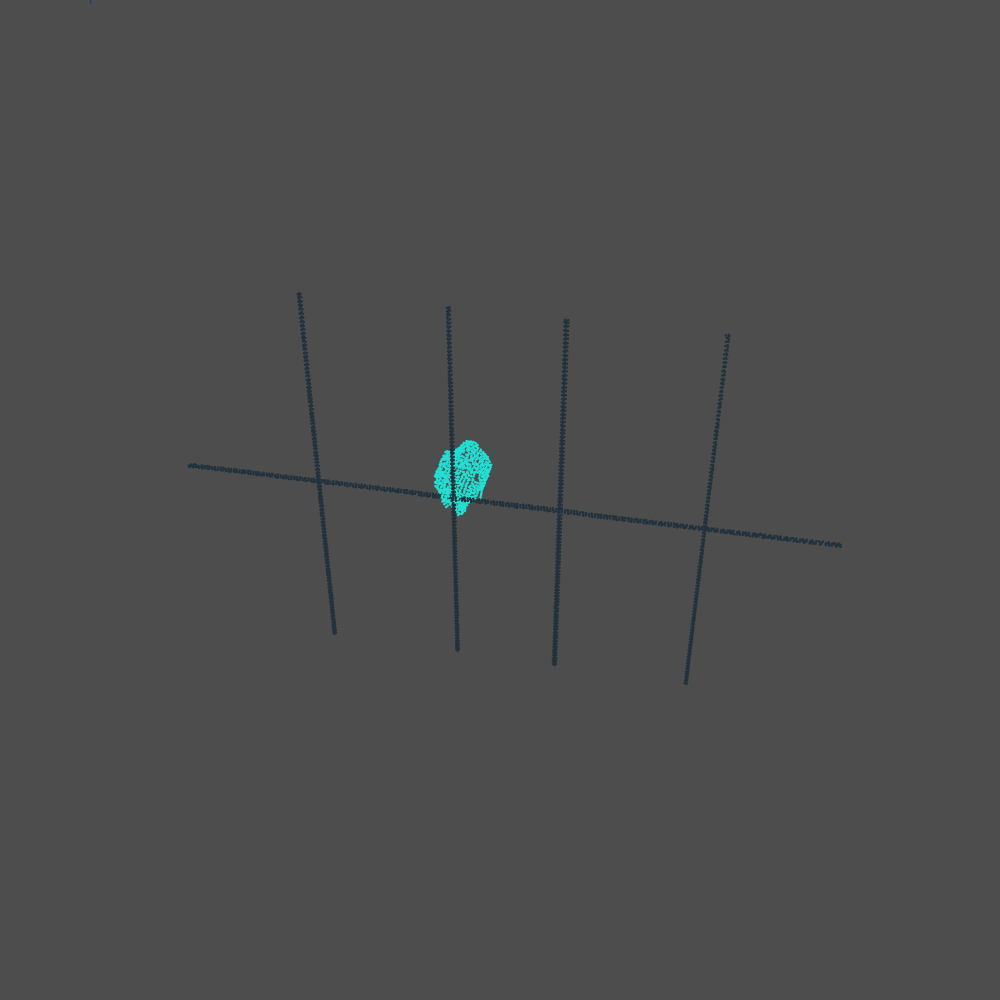}%
        \label{fig:demo1_step2}
    }
    \hspace{0.01\columnwidth}
    \subfloat[8-node demo, on Test \Romannum{4}]{
        \includegraphics[width=0.26\columnwidth]{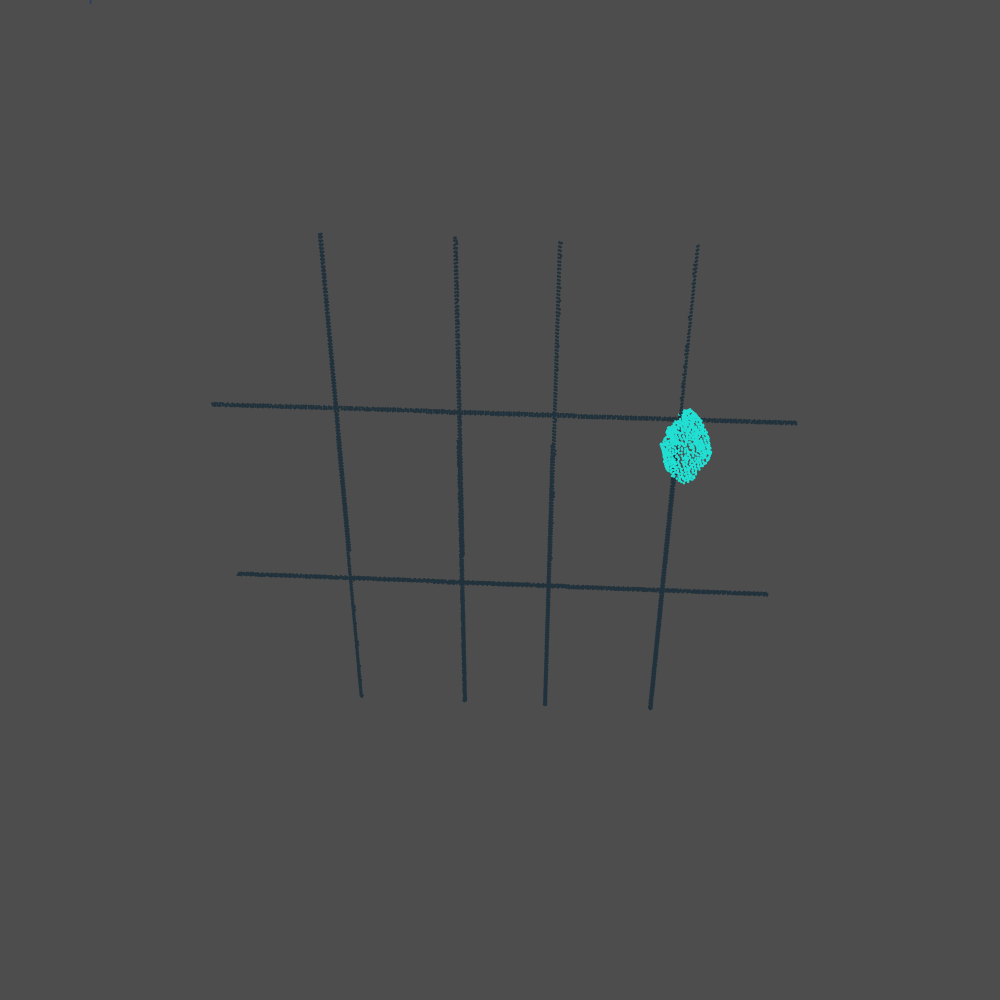}%
        \label{fig:demo1_step3}
    }

    %------------------ 换行 ------------------%

    %------------------ 第二行 ------------------%
    \subfloat[16-node demo, on Test \Romannum{7}]{
        \includegraphics[width=0.26\columnwidth]{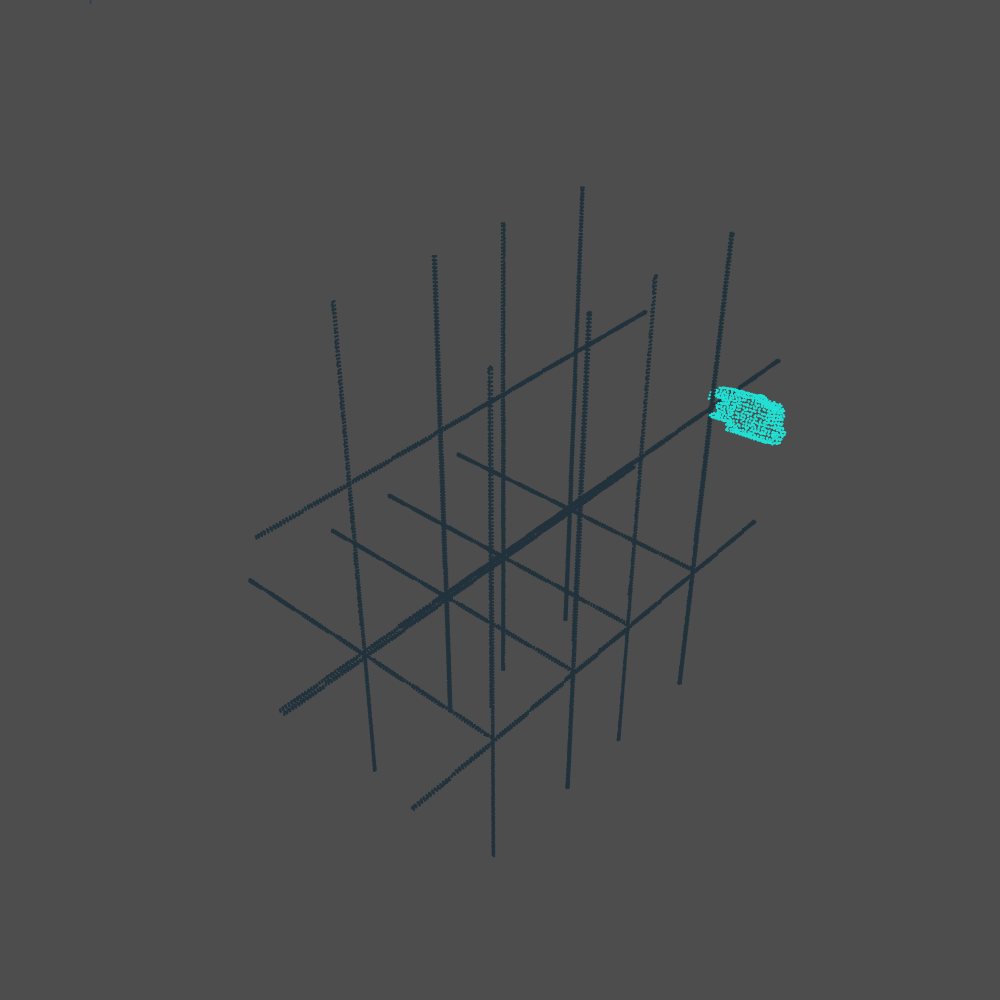}%
        \label{fig:demo2_step1}
    }
    \hspace{0.01\columnwidth}
    \subfloat[4-node demo with noise, on Test \Romannum{3}]{
        \includegraphics[width=0.26\columnwidth]{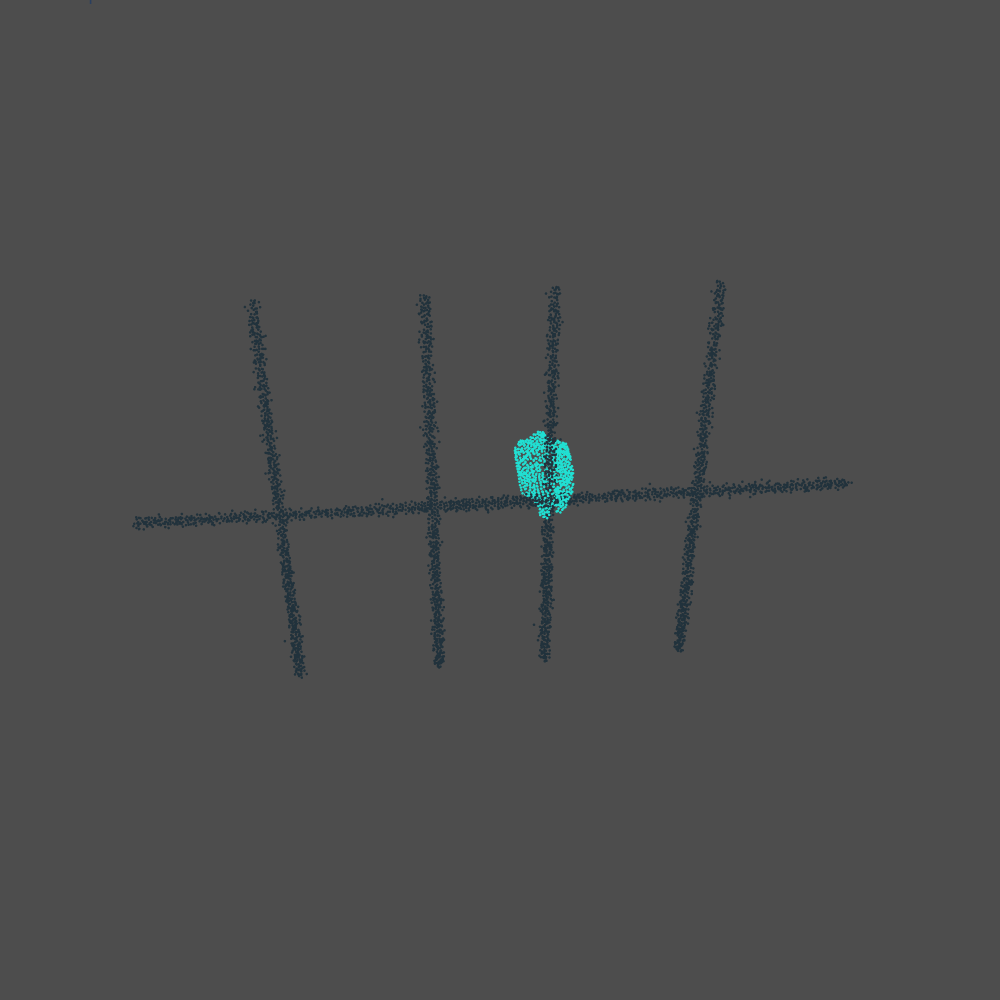}%
        \label{fig:demo2_step2}
    }
    \hspace{0.01\columnwidth}
    \subfloat[8-node demo with background,on Test \Romannum{5}]{
        \includegraphics[width=0.26\columnwidth]{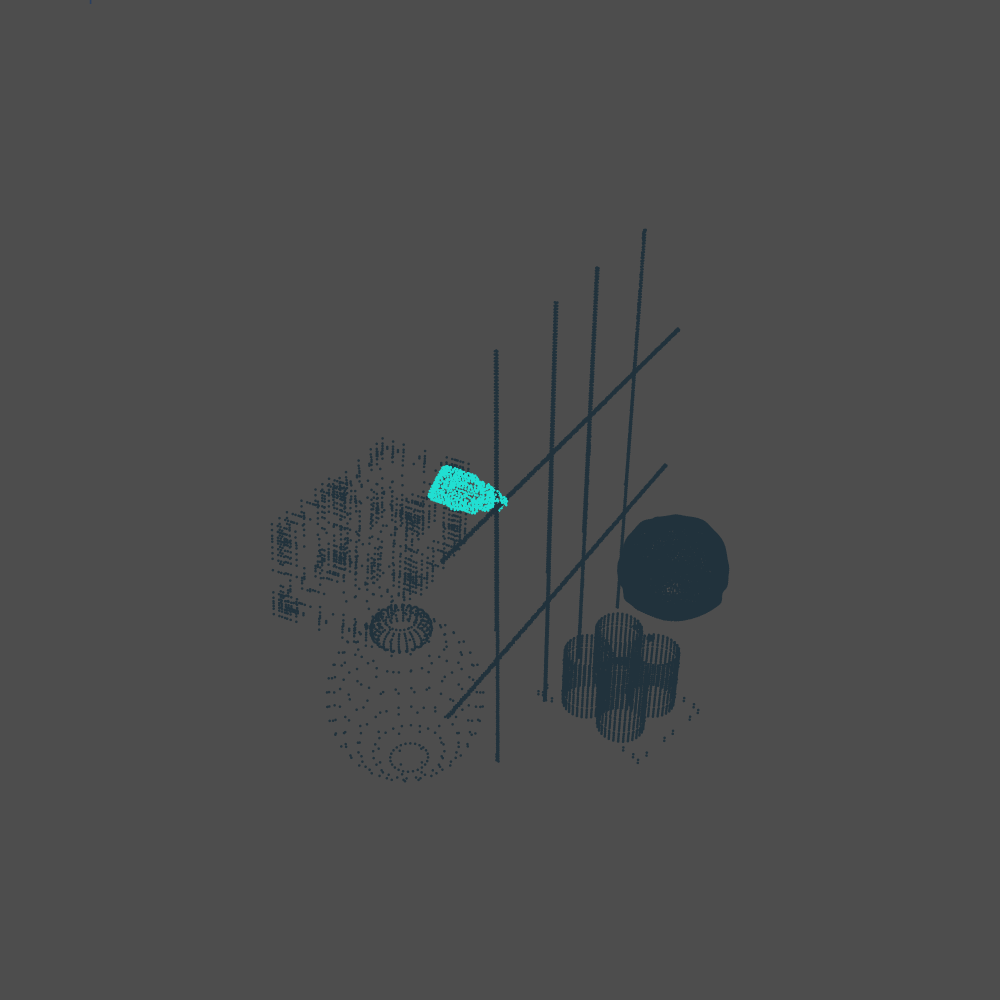}%
        \label{fig:demo2_step3}
    }

    \caption{%
        Visualizations of a subset of the datasets. (a) depicts a 1-node demo for train. (b)--(d) illustrate the rebar point cloud with 4 nodes, 8 nodes, 16 nodes. (e) shows the demo with noise and (f) shows the demo with 4 background obstacles.
        % Two visualizations of the tying processes by using \ac{cedf} with a tying gun as the end-effector.
        % The top row (a)--(c) illustrate the tying process of data from \textbf{Dataset B} (Gaussian noise in $[0, 0.5]$).
        % The bottom row (d)--(f) illustrate results on \textbf{Dataset C}, including 2 background obstacles and noise ($[0, 0.5]$).
        % (a) and (d) depict the initial pose,
        % (b) and (e) show the gripper tying the first tying node, and 
        % (c) and (f) show the gripper tying the second tying node.
    }
    \label{fig:demo_visual}
\end{figure}

The training sets (\textbf{Train \Romannum{1}} and \textbf{Train \Romannum{2}}) have 10 demostrations used to train the single-node detection model. \textbf{Train \Romannum{1}} is for the pre-detection part, and \textbf{Train \Romannum{2}} is for the later single node detection part. To assess the robustness of our algorithm, we focus on three key capabilities of the pipeline, e.g., resistance to noise, resilience against background obstacles, and performance with complex node distributions. For clarity, we measure the success rate based on the total number of detected nodes and compare these outcomes across datasets that vary in Gaussian noise levels, quantity of background obstacles, and intricate test scenarios. The success rate and prediction error to evaluate the algorithm in grasping tasks are calculated as follows:
\begin{itemize}
    \item The success rate ($R_s$) is calculated using the formula:
    \begin{equation}
    R_s = \frac{1}{N} \sum_{i=1}^{N} \frac{N_{si}}{L},
    \end{equation}
    where \( N_{si} \) represents the number of successful predictions at pose \( i \). A prediction is considered successful if
    \begin{equation}
    D_g(g, \hat{g}) < T_g,
    \end{equation}
    with $D_g(g, \hat{g})$ denoting the distance between the ground-truth $g$ and predicted poses $\hat{g}$, respectively, and \( T_g \) being the distance threshold for determining tying success.
    \item The prediction error $E_r$ is defined as the average pose distance between the predicted poses and the target ground-truth poses over all poses and demonstrations. It is calculated by using
    \begin{equation}
    E_r = \frac{1}{N} \sum_{i=1}^{N} \left[ \frac{1}{L} \sum_{l=1}^{L} D_g(g_{il}, \hat{g}_{il}) \right],
    \end{equation}
    where \( g_{il} \) and \( \hat{g}_{il} \) are the ground-truth and predicted poses for the \( i \)-th demonstration and the \( l \)-th pose, respectively. Herein we use the pose distance \( D_g \)
    to measure the difference between two poses \( g_1 = \{q_1, p_1\} \) and \( g_2 = \{q_2, p_2\} \) defined in quaternion form, which combines the angular distance and the linear distance. The pose distance is calculated as 
    \begin{equation}
    D_g(g_1, g_2) = d_p(p_1, p_2) + \gamma \theta_q(q_1, q_2),
    \end{equation}
    where
    \begin{equation}
    d_t(p_1, p_2) = \|p_1 - p_2\|^2,
    \end{equation}
    and
    \begin{equation}
    \theta_q(q_1, q_2) = 2 \cos^{-1}(\operatorname{real}(q_1 \cdot \bar{q_2})),
    \end{equation}
    with \( \gamma \) being the regularization term, \( \bar{q_2} \) the conjugate of \( q_2 \), and \( \gamma \) the regularization term.
\end{itemize}
We select three representative datasets to evaluate our algorithm, e.g., the first denotes rebar nodes on a line in \textbf{Test \Romannum{2}}, the second rebar nodes on a plane in \textbf{Test \Romannum{5}}, and the third denotes rebar nodes in space in \textbf{Test \Romannum{8}}. As our method works on SE(3) with 3D point cloud input, it is not necessary to compare it with the object detection-based pipeline. Thus, we test the model on diverse simulated data under different conditions and real-world data for verifying the performance of the pipeline.
% the first one featuring linear nodes, the second with planar nodes, and the third encompassing three-dimensional area 
% \textcolor{red}{?} 
% Text heads organize the topics on a relational, hierarchical basis. For example, the paper title is the primary text head because all subsequent material relates and elaborates on this one topic. If there are two or more sub-topics, the next level head (uppercase Roman numerals) should be used and, conversely, if there are not at least two sub-topics, then no subheads should be introduced. Styles named ÒHeading 1Ó, ÒHeading 2Ó, ÒHeading 3Ó, and ÒHeading 4Ó are prescribed.

\subsection{Experiment Results}
% The outcomes presented in Table \ref{tab:detection_res} demonstrate the effectiveness of our method across resistance to noise, resilience against background obstacles, and performance with complex node distributions.\textcolor{red}{what is areas}. Our approach successfully identifies rebar nodes, even in complex distributions with up to 36 nodes, achieving a success rate of 98\%. Despite the presence of noise, the success rates remain above 90\%. Even with significantly obvious background obstacles, our method maintains a success rate exceeding 80\%. 
The results in Table \ref{tab:node_detection_result} highlight the robustness of our method in handling noise, background obstacles, and complex node distributions. Our approach achieves a 98\% success rate in identifying rebar nodes, even in highly complex distributions with  36 nodes. It maintains success rates above 90\% in noisy environments and exceeds 80\% in scenarios with significant background interference.
% Positioning Figures and Tables: Place figures and tables at the top and bottom of columns. Avoid placing them in the middle of columns. Large figures and tables may span across both columns. Figure captions should be below the figures; table heads should appear above the tables. Insert figures and tables after they are cited in the text. Use the abbreviation ÒFig. 1Ó, even at the beginning of a sentence.

\begin{table}[h]
\caption{Nodes Detection Results}
\label{tab:node_detection_result}
\begin{center}

% \textbf{Robustness to noise} & 0 & 1-node,BG(2 Ob),noise(-)  \\
% \textbf{Robustness to background obstacles} & 50 & 4-node,BG(-),noise(-)  \\
% \textbf{Effectiveness for complex node distribution} & 50 & 8-node,BG(-),noise(-)  \\

\begin{tabular}{|p{45pt}|p{35pt}|p{100pt}|p{20pt}|}
\hline
Evaluation purpose & Symbol & Condition & $R_s$ \\
\hline
\multirow{9}{50pt}{\textbf{Robustness to noise}} & \textbf{Test \Romannum{1}} & 4-node,BG(-), noise(-) & $96\%$ \\
\ & \textbf{Test \Romannum{3}} & 4-node, BG (-), noise ([0,0.5]) & $98\%$\\
\cline{2-4}
\ & \textbf{Test \Romannum{4}} & 8-node, BG (-), noise (-) & $94\%$\\
\ & \textbf{Test \Romannum{6}} & 8-node, BG (-), noise ([0,0.5]) & $90\%$ \\
\cline{2-4}
\ & \textbf{Test \Romannum{7}} & 16-node, BG (-), noise (-) & $98\%$\\
\ & \textbf{Test \Romannum{9}} & 16-node, BG (-), noise ([0,0.5]) &$94\%$\\
\hline
\multirow{9}{50pt}{\textbf{Robustness to background obstacles}} & \textbf{Test \Romannum{1}} & 4-node, BG (-), noise (-) & $96\%$\\
\ & \textbf{Test \Romannum{2}} & 4-node, BG (4 Ob), noise (-) & $90\%$\\
\cline{2-4}
\ & \textbf{Test \Romannum{4}} & 8-node, BG (-), noise (-) & $94\%$\\
\ & \textbf{Test \Romannum{5}} & 8-node, BG (4 Ob), noise (-) & $82\%$\\
\cline{2-4}
\ & \textbf{Test \Romannum{7}} & 16-node, BG (-), noise (-) & $98\%$\\
\ & \textbf{Test \Romannum{8}} & 16-node, BG (4 Ob), noise (-) & $84\%$\\
% \hline
% - & reference samples & 10 & \multirow{6}{*}{training} \\
% - & optimizer & Adam &  \\
% $lr$ & learning rate  & 0.0003 &  \\
% $\beta$ & momentum    & [0.9, 0.98] &  \\
% $\epsilon$ & epsilon  & $1\times 10^{-9}$ &  \\
% - & weight decay         & 0.0001 &  \\
\hline
\multirow{1}{50pt}{\textbf{Effectiveness for complex node distribution}} & \textbf{Test \Romannum{1}} & 4-node, BG (-), noise (-) & $96\%$\\
\ & \textbf{Test \Romannum{4}} & 8-node, BG (-), noise (-) & $94\%$\\
\ & \textbf{Test \Romannum{7}} & 16-node, BG (-), noise (-) & $98\%$\\
\ & \textbf{Test \Romannum{10}} & 32-node, BG (-), noise (-) & $96\%$\\
\ & \textbf{Test \Romannum{11}} & 36-node, BG (-), noise (-) & $98\%$\\
\hline
\multicolumn{4}{p{225pt}}{BG: background, noise: Gaussian noise, obstacles: Ob} \\
\multicolumn{4}{p{225pt}}{-: not considered (clear)}
\end{tabular}%
\end{center}
\end{table}

%    \begin{figure}[thpb]
%       \centering
%       \framebox{\parbox{3in}{We suggest that you use a text box to insert a graphic (which is ideally a 300 dpi TIFF or\textbf{Eps}file, with all fonts embedded) because, in an document, this method is somewhat more stable than directly inserting a picture.
% }}
%       %\includegraphics[scale=1.0]{figurefile}
%       \caption{Inductance of oscillation winding on amorphous
%        magnetic core versus DC bias magnetic field}
%       \label{figurelabel}
%    \end{figure}
   
\begin{figure}[!t]
    \centering
    %------------------ 第一行 ------------------%
    \subfloat[The 36-node demo, on \textbf{Train \Romannum{11}}]{
        \includegraphics[width=0.35\columnwidth]{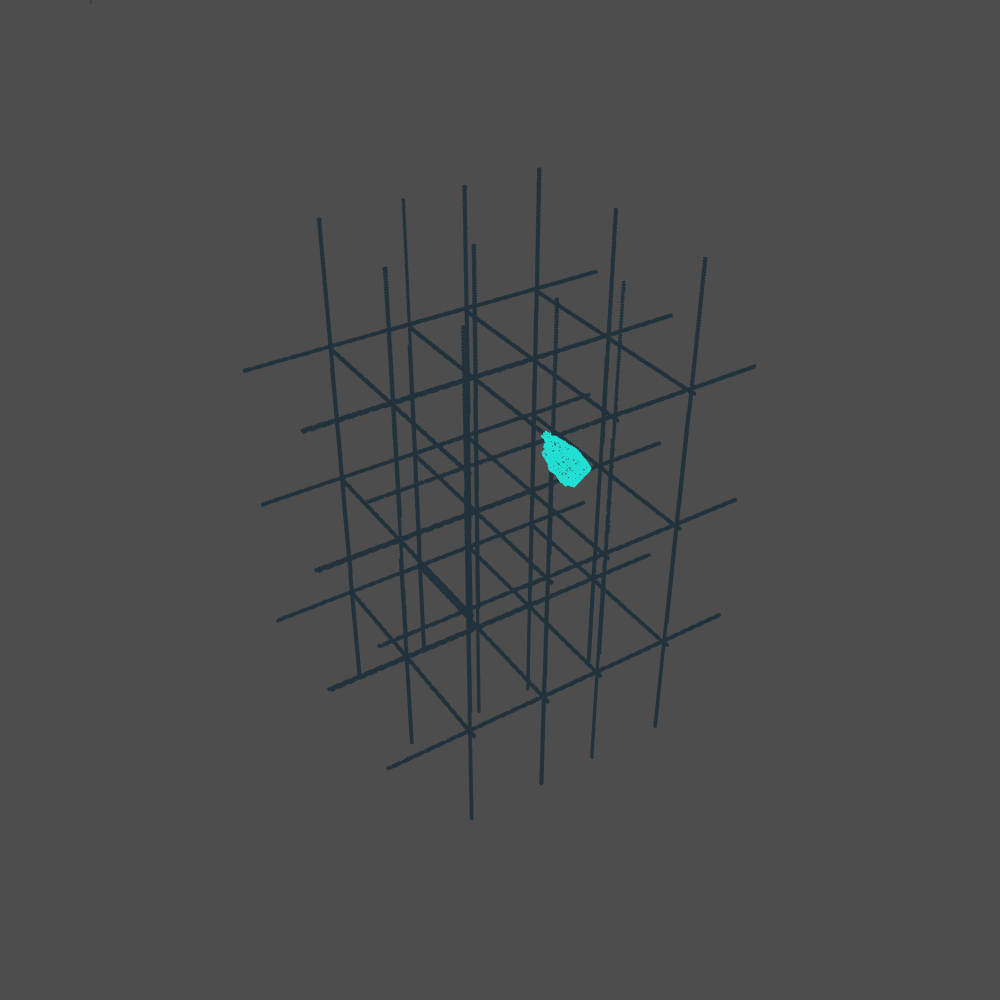}%
        \label{fig:demo1_step1}
    }
    \hspace{0.01\columnwidth}
    \subfloat[The nodes and direction vectors of the 36-node demo in (a), on \textbf{Test \Romannum{11}}]{
        \includegraphics[width=0.38\columnwidth]{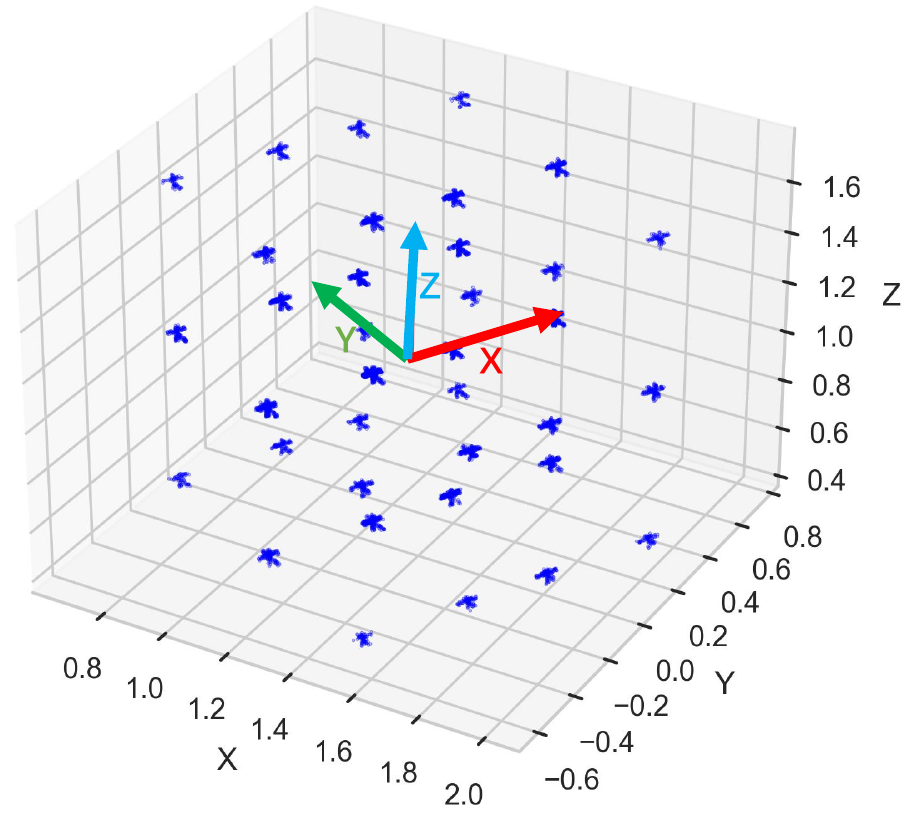}%
        \label{fig:demo1_step2}
    }

    %------------------ 换行 ------------------%
    \vspace{0em}
    %------------------ 第二行 ------------------%
    \subfloat[The 32-node demo, on \textbf{Test \Romannum{10}}]{
        \includegraphics[width=0.35\columnwidth]{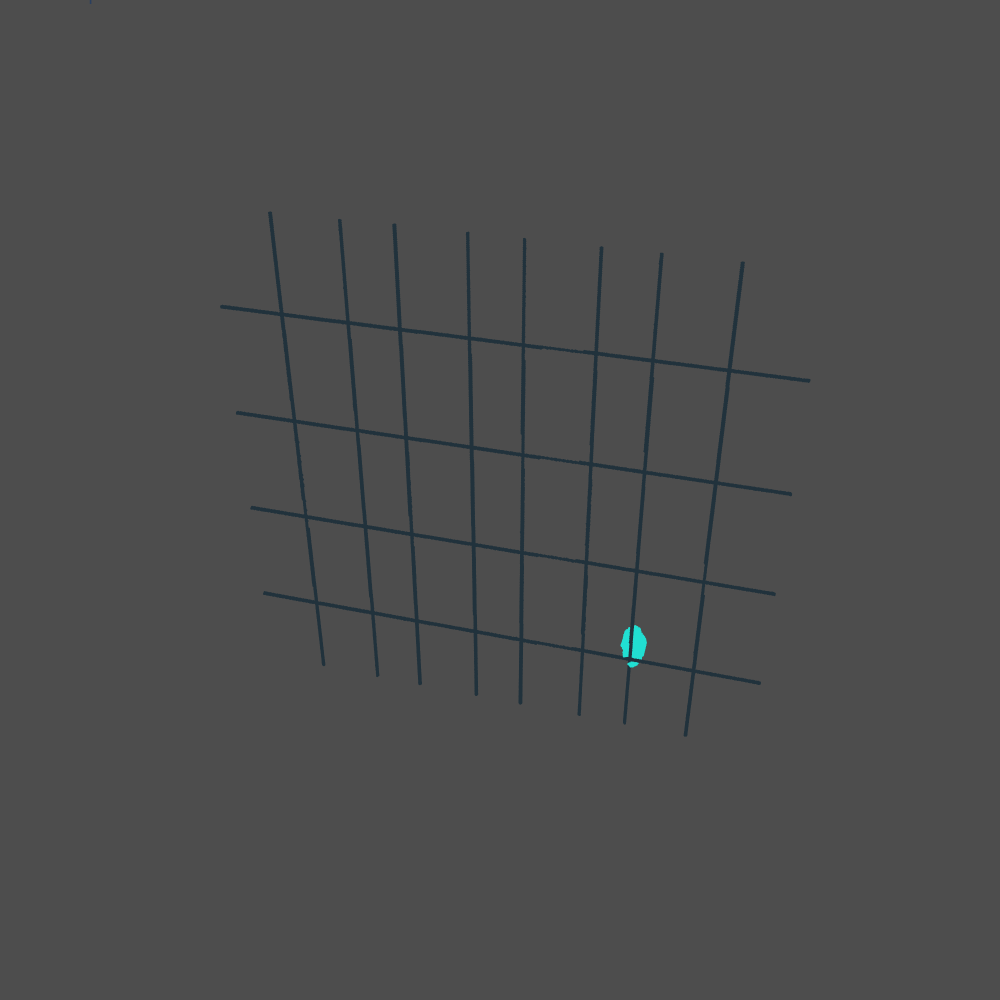}%
        \label{fig:demo2_step1}
    }
    \hspace{0.01\columnwidth}
    \subfloat[The nodes and direction vectors of the 32-node demo in (c), on \textbf{Test \Romannum{10}}]{
        \includegraphics[width=0.38\columnwidth]{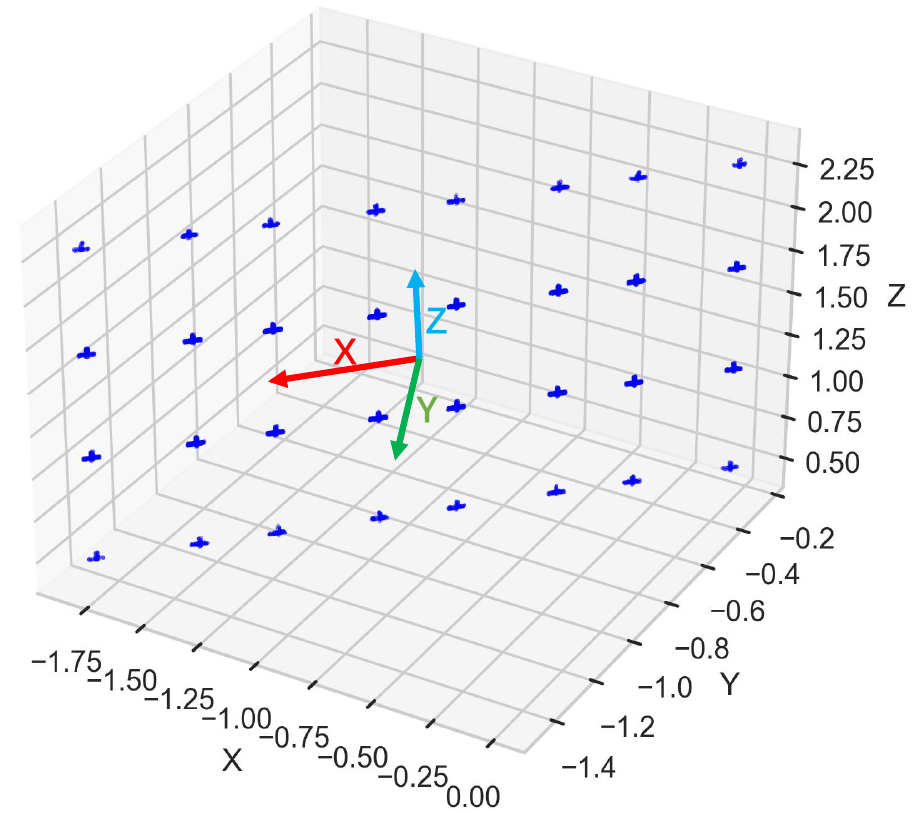}%
        \label{fig:demo2_step2}
    }

     %------------------ 换行 ------------------%
    \vspace{0em}
    %------------------ 第三行 ------------------%
    \subfloat[The 16-node demo, on \textbf{Test \Romannum{8}}]{
        \includegraphics[width=0.35\columnwidth]{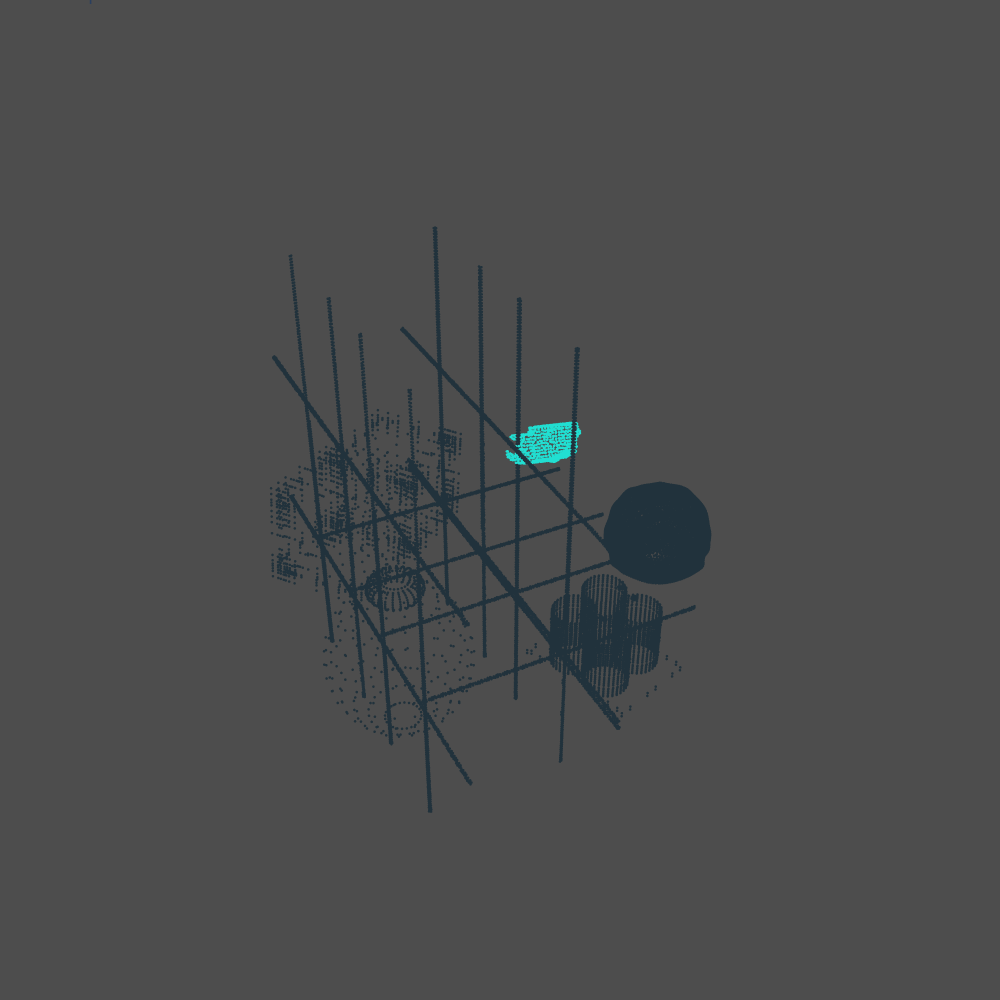}%
        \label{fig:demo2_step1}
    }
    \hspace{0.01\columnwidth}
    \subfloat[The nodes and direction vectors of the 16-node demo in (e), on \textbf{Test \Romannum{8}}]{
        \includegraphics[width=0.38\columnwidth]{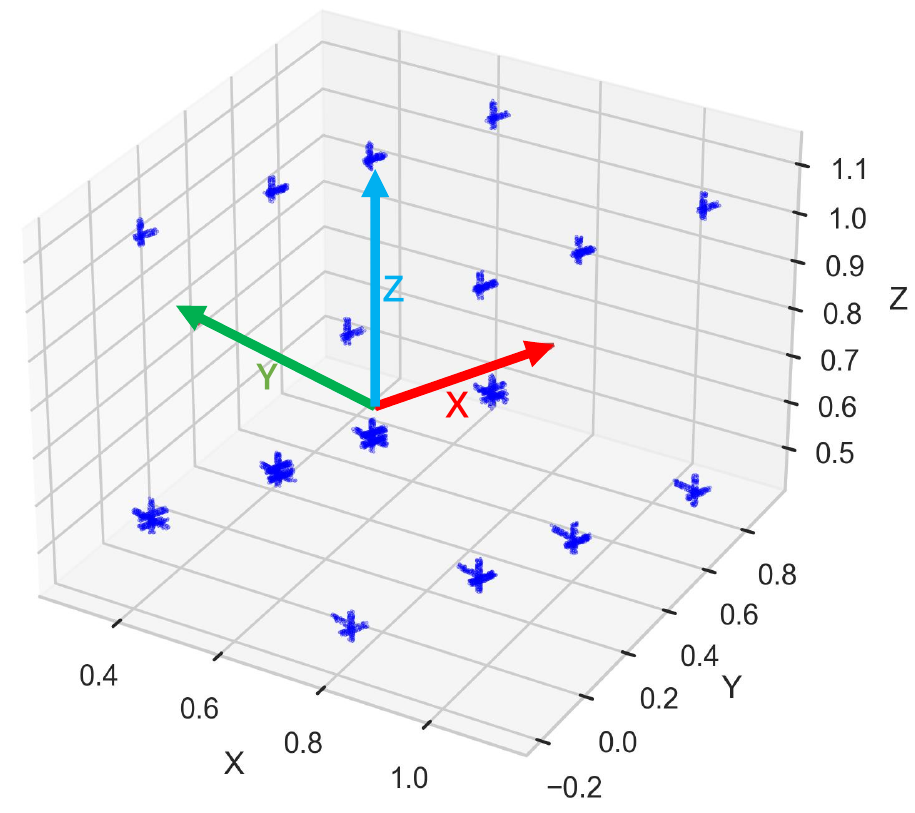}%
        \label{fig:demo2_step2}
    }

     %------------------ 换行 ------------------%
    \vspace{0em}
    %------------------ 第三行 ------------------%
    \subfloat[The 8-node demo with noise, on \textbf{Test \Romannum{6}}]{
        \includegraphics[width=0.35\columnwidth]{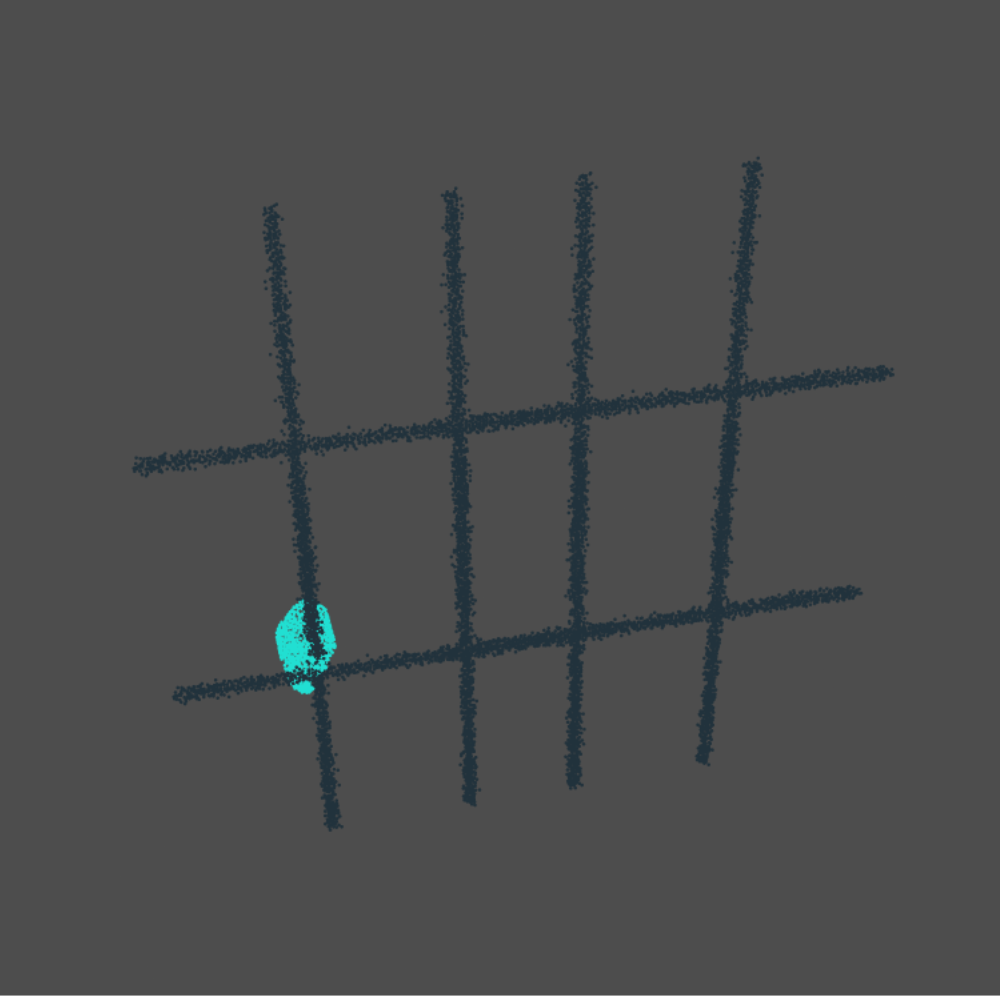}%
        \label{fig:demo2_step1}
    }
    \hspace{0.01\columnwidth}
    \subfloat[The nodes and direction vectors of the 8-node demo in (g), on \textbf{Test \Romannum{6}}]{
        \includegraphics[width=0.38\columnwidth]{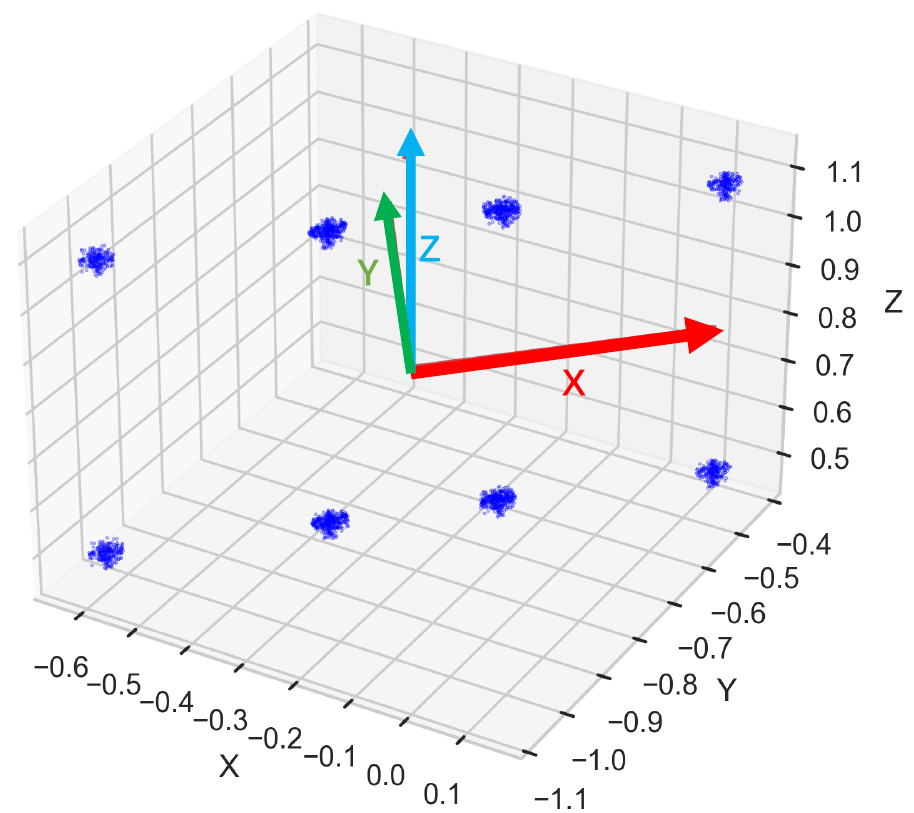}%
        \label{fig:demo2_step2}
    }

    \caption{%
        Visualizations of node detection in \textbf{Test \Romannum{11}}, \textbf{Test \Romannum{10}}, \textbf{Test \Romannum{8}} and \textbf{Test \Romannum{6}}. (a) shows the 36-node demonstration and (b) depicts the extracted nodes and the estimation of their direction vectors in \textbf{Test \Romannum{11}}. (c) and (d) illustrate the results in the 32-node demonstration in in \textbf{Test \Romannum{10}}. (e) and (f) illustrate the results in the 16-node demonstration scenario with background obstacles in \textbf{Test \Romannum{8}}. (g) and (h) illustrate the results in the 8-node demonstration scenario with Gaussian noise in \textbf{Test \Romannum{6}}.
    }
    \label{fig:nodesresult_visualization}
\end{figure}

\begin{figure}[!t]
    \centering
    %------------------ 第一行 ------------------%
    \subfloat[$E_r$ of \textbf{Test \Romannum{2}}, \textbf{Test \Romannum{5}}, \textbf{Test \Romannum{8}} ]{
        \includegraphics[width=0.72\columnwidth]{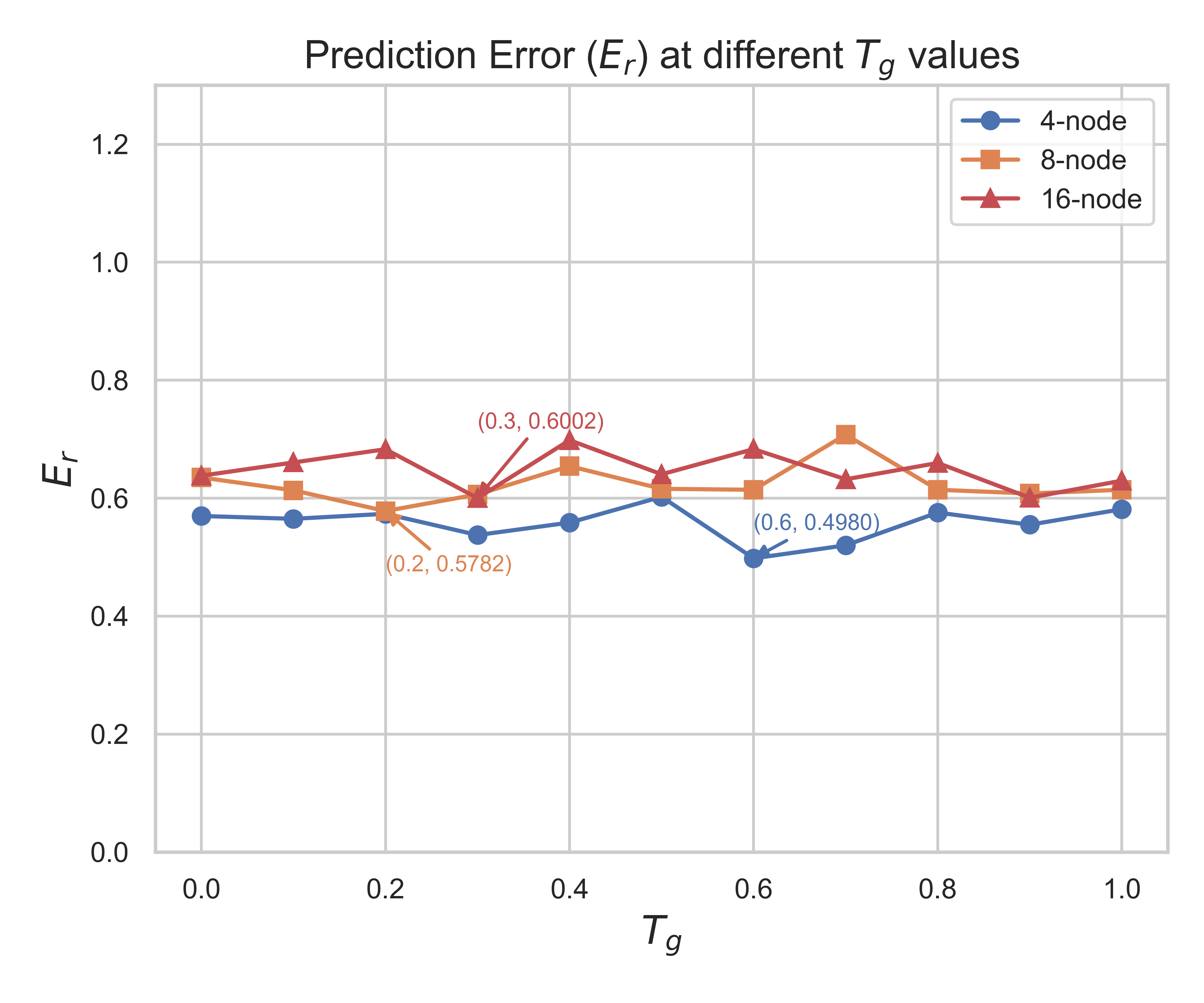}%
        \label{fig:Er}
    }
     %------------------ 换行 ------------------%
    \vspace{0em}
    
    \subfloat[$R_s$ of \textbf{Test \Romannum{2}}, \textbf{Test \Romannum{5}}, \textbf{Test \Romannum{8}} ]{
        \includegraphics[width=0.8\columnwidth]{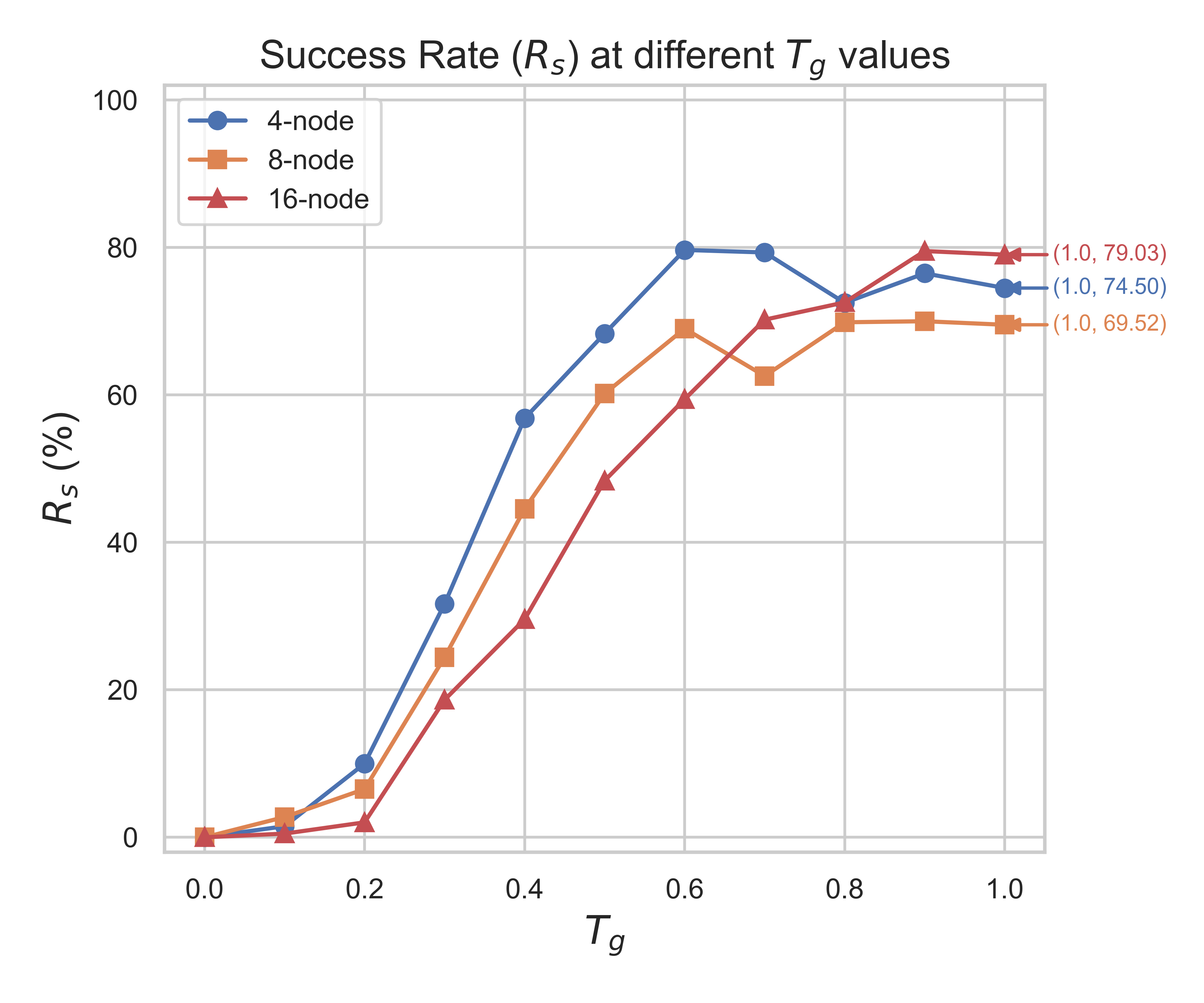}%
        \label{fig:Rs}
    }
    \caption{%
       (a) and (b) respectively illustrate the prediction error ($E_r$) and success rate ($R_s$) for 4-node datasets in \textbf{Test \Romannum{2}}, 8-node datasets in \textbf{Test \Romannum{5}}, and 16-node datasets in \textbf{Test \Romannum{8}}, under conditions of no noise and 4 background obstacles. 
    }
    \label{fig:evaluate}
\end{figure}

% Figure Labels: Use 8 point Times New Roman for Figure labels. Use words rather than symbols or abbreviations when writing Figure axis labels to avoid confusing the reader. As an example, write the quantity ÒMagnetizationÓ, or ÒMagnetization, MÓ, not just ÒMÓ. If including units in the label, present them within parentheses. Do not label axes only with units. In the example, write ÒMagnetization (A/m)Ó or ÒMagnetization {A[m(1)]}Ó, not just ÒA/mÓ. Do not label axes with a ratio of quantities and units. For example, write ÒTemperature (K)Ó, not ÒTemperature/K.Ó

Furthermore, Fig. \ref{fig:nodesresult_visualization} demonstrates the precision of our method in detecting rebar nodes and estimating direction vectors, even in highly complex configurations such as the 32-node and 36-node setups. Our filtering technique effectively minimizes background interference, ensuring accurate node data extraction even when background elements are in close proximity to the rebar point cloud. Additionally, the results highlight the robustness of our method to random Gaussian noise.

Fig. \ref{fig:evaluate} presents a comprehensive comparison of the prediction error ($E_r$) and success rate ($R_s$) across three datasets: 4-node datasets in \textbf{Test \Romannum{2}}, 8-node datasets in \textbf{Test \Romannum{5}}, and 16-node datasets in \textbf{Test \Romannum{8}}, evaluated under conditions of no noise and 4 background obstacles, with varying threshold values ($T_g$).
Fig. \ref{fig:Er} illustrates the prediction error ($E_r$), showing that the error remains relatively stable across different $T_g$ values, with only minor fluctuations. Notably, the 4-node dataset in \textbf{Test \Romannum{2}} exhibits the lowest prediction error, closely followed by the 8-node dataset in \textbf{Test \Romannum{5}} and the 16-node dataset in \textbf{Test \Romannum{8}}, all maintaining errors around 0.6.

Fig. \ref{fig:Rs} illustrates the success rate ($R_s$) for the same datasets shown in Fig. \ref{fig:Er}, evaluated across different $T_g$ values. The success rate increases significantly as $T_g$ approaches 1. The 16-node dataset in \textbf{Test \Romannum{8}} achieves the highest success rate, peaking at 79.63\% at $T_g$=1, whereas the 4-node dataset in \textbf{Test \Romannum{2}} and the 8-node dataset in \textbf{Test \Romannum{5}} show slightly lower success rates of 74.50\% and 69.52\%, respectively.
These results underscore the robustness and reliability of our method across different node configurations, highlighting its potential for practical applications in real-world scenarios. The consistent performance across varying $T_g$ values demonstrates the method's effectiveness in accurately predicting rebar nodes and estimating direction vectors, even in complex distributions.

\subsection{Real-world Test and Limitations}
\begin{figure}[!t]
    \centering
    %------------------ 第一行 ------------------%
    %\subfloat[Starting point]{
    %    \includegraphics[width=0.29\columnwidth]{pic/real5_first.png}%
    %    \label{fig:real5_first}
    %}
    \subfloat[Node 1]{
        \includegraphics[width=0.24\columnwidth]{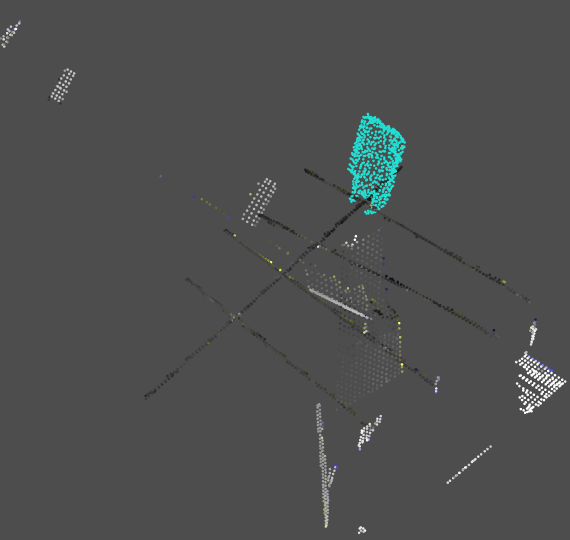}%
        \label{fig:real4_1}
    }
    \subfloat[Node 2]{
        \includegraphics[width=0.24\columnwidth]{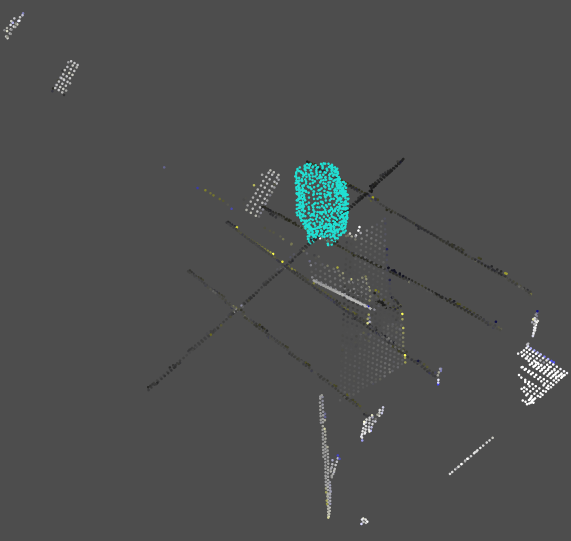}%
        \label{fig:real4_2}
    }
    \subfloat[Node 3]{
        \includegraphics[width=0.24\columnwidth]{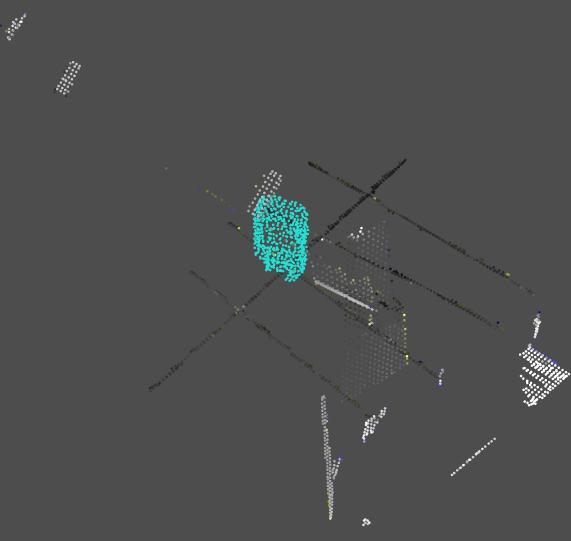}%
        \label{fig:real4_3}
    }
    \subfloat[Node 4]{
        \includegraphics[width=0.24\columnwidth]{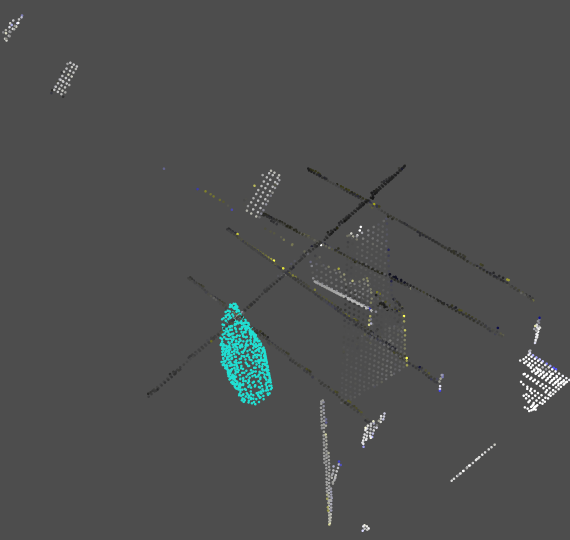}%
        \label{fig:real4_4}
    }
    %\subfloat[Node 5]{
    %    \includegraphics[width=0.29\columnwidth]{pic/real5_5.png}%
    %    \label{fig:real5_5}
    %}
    \caption{%
        Visualizations of tying process on real data with 4 nodes.
        % Two visualizations of the tying processes by using \ac{cedf} with a tying gun as the end-effector.
        % The top row (a)--(c) illustrate the tying process of data from \textbf{Dataset B} (Gaussian noise in $[0, 0.5]$).
        % The bottom row (d)--(f) illustrate results on \textbf{Dataset C}, including 2 background obstacles and noise ($[0, 0.5]$).
        % (a) and (d) depict the initial pose,
        % (b) and (e) show the gripper tying the first tying node, and 
        % (c) and (f) show the gripper tying the second tying node.
    }
    \label{fig:real5}
\end{figure}

\begin{figure}[!t]
    \centering
    %------------------ 第一行 ------------------%
    \subfloat[Failure of generating tying pose]{
        \includegraphics[width=0.3\columnwidth]{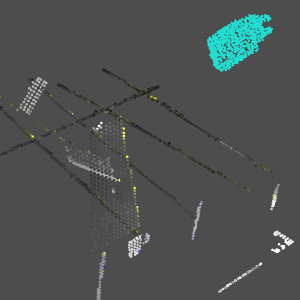}%
        \label{fig:failure1_not_grasp}
    }
    \hspace{0.01\columnwidth}
    \subfloat[Generation of wrong tying pose]{
        \includegraphics[width=0.3\columnwidth]{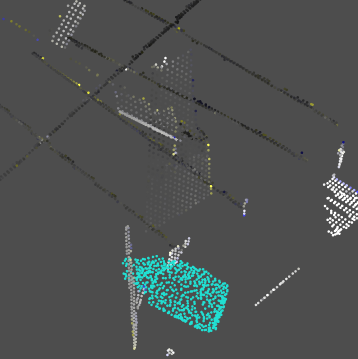}%
        \label{fig:failure2_incorrect_grasp}
    }

    \caption{%
        Visualizations of tying failure.
        % Two visualizations of the tying processes by using \ac{cedf} with a tying gun as the end-effector.
        % The top row (a)--(c) illustrate the tying process of data from \textbf{Dataset B} (Gaussian noise in $[0, 0.5]$).
        % The bottom row (d)--(f) illustrate results on \textbf{Dataset C}, including 2 background obstacles and noise ($[0, 0.5]$).
        % (a) and (d) depict the initial pose,
        % (b) and (e) show the gripper tying the first tying node, and 
        % (c) and (f) show the gripper tying the second tying node.
    }
    \label{fig:demo_visual}
\end{figure}

To verify the potential of our system in a real-world application, we use our hardware system (details in Section \ref{sec:method}) to collect real-world data with 4 nodes. The model trained on simulated data is applied directly on the collected real data, and the tying process is visualized in Fig. \ref{fig:real5}. Although the reabrs are positioned in an uneven manner and with noisy background, the pipeline can still achieve the sequential 4-node tying. However, some failures can also be observed, e.g., failure of generating a tying pose (Fig. \ref{fig:failure1_not_grasp}) and generation of a wrong tying pose (Fig. \ref{fig:failure2_incorrect_grasp}). These limitations can possibly be resolved by training the model on complex real-world datasets.

\section{CONCLUSIONS}
% This paper introduces a multi-node retrieval architecture designed to address rebar tying tasks in industrial settings. Utilizing the Diffusion-EDFs model for single node detection, coupled with an effective filtering technique for rebar node extraction and orderly ranking through PCA-based direction vector estimation, our approach demonstrates robust performance in our tests. Future work may involve applying our methods to real-world scenarios and further enhancing model detection accuracy.
In this paper, we propose an integrated framework for rebar node detection and sorting tailored to industrial rebar tying tasks. Our system features a single-node detection module based on a diffusion generative model in SE(3) and an adaptive orthogonal feature filtering algorithm that leverages the geometric characteristics of rebar structures. Combined with a coordinate system established via a principal component analysis (PCA), this approach enables efficient extraction and accurate sorting of rebar nodes. Unlike traditional methods, our framework demonstrates exceptional robustness in single-node detection and stable node recognition under complex and noisy conditions through multi-level feature fusion and effective spatial information utilization. Experimental results show that our method effectively handles noise, background obstacles, and complex node configurations, robustly supporting the positioning and grasping tasks of industrial robots in rebar tying. In the future work, we will further optimize model parameters and explore deploying this framework in practical industrial scenarios, continuously improving detection and positioning accuracy to better support automation in related fields.
% A conclusion section is not required. Although a conclusion may review the main points of the paper, do not replicate the abstract as the conclusion. A conclusion might elaborate on the importance of the work or suggest applications and extensions. 

\addtolength{\textheight}{-12cm}   % This command serves to balance the column lengths

%%%%%%%%%%%%%%%%%%%%%%%%%%%%%%%%%%%%%%%%%%%%%%%%%%%%%%%%%%%%%%%%%%%%%%%%%%%%%%%%

%%%%%%%%%%%%%%%%%%%%%%%%%%%%%%%%%%%%%%%%%%%%%%%%%%%%%%%%%%%%%%%%%%%%%%%%%%%%%%%%



\section*{ACKNOWLEDGMENT}

This paper was funded by InnoHK-HKCRC and Shenzhen Startup Funding (No. QD2023014C) supported by Meituan.

% %%%%%%%%%%%%%%%%%%%%%%%%%%%%%%%%%%%%%%%%%%%%%%%%%%%%%%%%%%%%%%%%%%%%%%%%%%%%%%%%

% References are important to the reader; therefore, each citation must be complete and correct. If at all possible, references should be commonly available publications.

\end{document}